\documentclass{article}
\usepackage{arxiv}

\title{Evaluation-Free Recursive Self-Improvement for Coding Agents}

\author{%
  \normalfont \textbf{Zhe Ren}\textsuperscript{1}, \textbf{Guowei Rong}\textsuperscript{1},
  \textbf{Yimeng Chen}\textsuperscript{2}, \textbf{Li Nanbo}\textsuperscript{2}\\[3pt]
  \normalfont \textbf{Piotr Pi\k{e}kos}\textsuperscript{2},
  \textbf{Mateusz Ostaszewski}\textsuperscript{2}, \textbf{Dandan Guo}\textsuperscript{1,2,}\thanks{Corresponding author: \texttt{guodandan@jlu.edu.cn}.},
  \textbf{J\"urgen Schmidhuber}\textsuperscript{2}\\[8pt]
  \normalfont\small\textsuperscript{1}School of Artificial Intelligence, Jilin University\\[2pt]
  \normalfont\small\textsuperscript{2}King Abdullah University of Science and Technology (KAUST)%
}

\date{\today}
\renewcommand{\undertitle}{}

\renewcommand{\shorttitle}{Evaluation-Free Recursive Self-Improvement}
\hypersetup{
  pdftitle={Evaluation-Free Recursive Self-Improvement for Coding Agents},
  pdfauthor={Zhe Ren, Guowei Rong, Yimeng Chen, Li Nanbo, Piotr Piękos, Mateusz Ostaszewski, Dandan Guo, Jürgen Schmidhuber},
  pdfsubject={Evaluation-free recursive self-improvement for coding agents},
  pdfkeywords={coding agents, self-improvement, explore-then-integrate}
}

\begin{document}
\raggedbottom
\maketitle

\begin{abstract}
Recent self-improving coding agents have achieved strong benchmark performance by iteratively modifying their own executable implementations. A common pattern uses downstream task-derived evidence to decide which candidates or lineages continue evolving, which we term \emph{evaluate-and-select}. Obtaining this signal requires repeatedly executing evolved agents on downstream tasks, so broader exploration incurs increasing evaluation cost, while repeated optimization against a fixed evaluation distribution may encourage benchmark-specific adaptation. We ask a more fundamental question: \emph{can coding agents self-improve without downstream task evaluation inside the evolution loop?} We introduce \textbf{E}xp\textbf{l}ore-then-\textbf{I}n\textbf{t}egrat\textbf{e} (\method), which integrates implementation mechanisms rather than selecting a single exploratory child to continue the lineage. At each generation, \method proposes diverse focused improvements, implements them independently from the same parent, and inspects the resulting programs to identify complementary, redundant, and conflicting mechanisms before reimplementing a compatible subset as one coherent descendant. No downstream task outcomes or task-solving trajectories of evolved candidates are used to guide evolution; only deterministic checks of structural and operational validity are used inside the loop. On \textit{SWE-bench Pro}, \method improves the initial agent from 60.0\% to 85.6\%, compared with 86.7\% for the Darwin Gödel Machine (DGM) and 88.3\% for the Huxley-Gödel Machine (HGM), while using zero candidate task evaluations and reducing observed wall-clock evolution time by 8.8x and 3.2x, respectively, under the evaluated configurations. In cross-benchmark transfer without further adaptation, \method averages 73.3\% on \textit{SWE-bench Verified} and 69.3\% on \textit{Polyglot}. Together, these results demonstrate the feasibility of coding-agent self-improvement without downstream candidate task evaluation in our setting.
\end{abstract}
\section{Introduction}

The idea of AI systems improving their capabilities has a long history, from early informal discussions of machines designing more capable successors \citep{good1966speculations} to concrete algorithms for recursive self-improvement (RSI) \citep{schmidhuber1987selfreferential}
and the Gödel machine \citep{schmidhuber2003godel}. Recent work continues to pursue this goal through modern agentic systems \citep{fang2025comprehensivesurveyselfevolvingai, gao2025survey, ren2026self}. These systems usually improve either the underlying large language model (LLM) or its scaffold, where the latter comprises prompts \citep{fernando2023promptbreeder, yuksekgonul2024textgradautomaticdifferentiationtext}, memory \citep{xu2026mem, zhang2026memrl}, tools \citep{cai2024large, qiu2025alita}, and control logic \citep{zelikman2023self,yin2025godel} that govern how the model is used. Coding agents provide a practical setting for studying self-improvement because they can access and modify their own source code to recursively generate better variants of themselves. Recent systems have improved coding performance by repeatedly modifying this software while keeping the LLM fixed \citep{hu2025automated, zhang2025aflow, robeyns2025self, zhang2026darwin, wang2026huxley}. Since self-improvement can revisit agents retained in an archive,
a central design question is not only how to generate self-modifications, but also which modifications should be carried forward during subsequent search.

\begin{figure}[htbp]
    \centering
    \includegraphics[width=\linewidth]{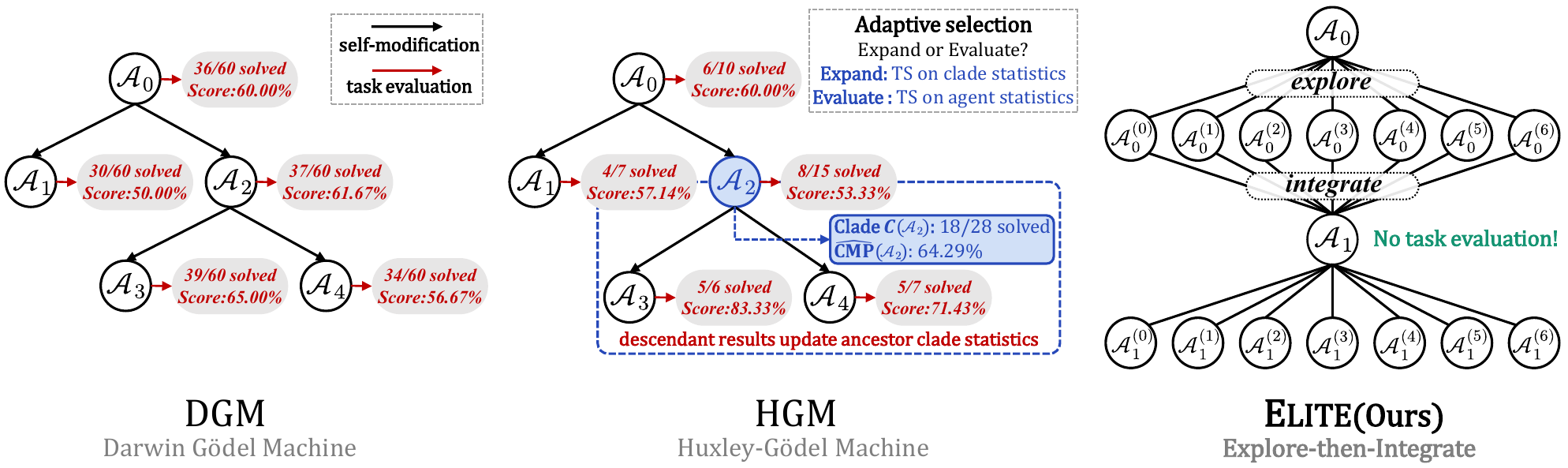}
    \caption{DGM and HGM exemplify \emph{evaluate-and-select} strategies, using task-derived evaluation to guide evolution. \method{} instead explores multiple same-parent implementations and integrates compatible mechanisms into a descendant, without using downstream task outcomes to guide evolution.}
    \label{fig:explore_synthesize}
\end{figure}

Existing approaches answer this question in different ways, but most rely on a common principle that we call \emph{evaluate-and-select} to decide which candidates or lineages continue evolving. Although their search and selection mechanisms differ, downstream task-derived evidence is used to determine which candidates, groups, or lineages receive subsequent evolutionary effort. DGM uses benchmark performance to guide
parent selection from an archive of agents \citep{zhang2026darwin}. HGM uses descendant task performance to estimate a lineage's potential for further improvement \citep{wang2026huxley}. GEA uses group-level search and task-based novelty
\citep{weng2026group}, while MGM shares information across lineages
\citep{liu2026mendelgodelmachinerecursive}. SIFT uses LLM-based comparisons to rank candidate patches \citep{fu2026selfimprovementfasttreesearch}, while RQGM evolves evaluation criteria during search \citep{iacob2026redqueengodelmachine}. This evaluate-and-select paradigm introduces two  limitations. First, obtaining reliable downstream feedback requires repeatedly executing evolved agents on target tasks, making broader exploration increasingly expensive \citep{xu2026verifysmarterevolvefurther, fu2026selfimprovementfasttreesearch, wang2026rethinkingevaluationharnessevolution, zeng2026dockerlessenvironmentfreeprogramverifier}. Second, candidate-level performance scores alone do not specify which implementation mechanisms should be retained or combined. Repeated optimization against a fixed evaluation distribution may further encourage benchmark-specific adaptation rather than transferable improvements \citep{kapoor2024ai, xia2026rrsiregularizedrecursiveselfimprovement}. We therefore ask: \textit{can coding agents decide which changes to carry forward without downstream candidate task evaluation}?

A key insight is that executable exploration can generate implementation-level evidence for inheritance decisions without downstream candidate task feedback. Implementing focused modifications independently from the same parent makes their mechanisms, interfaces, and dependencies explicit and directly comparable. We can then reason about which mechanisms are complementary, redundant, or conflicting. This allows the next descendant to be constructed from compatible mechanisms without first determining which exploratory child performs best on downstream tasks. Exploratory children thus serve as intermediate implementations that inform integration, with their concrete changes providing
the basis for deciding what to inherit. This suggests a different route to recursive self-improvement: explore diverse modifications, inspect their implementations, and integrate compatible mechanisms without using downstream task performance to decide what is carried forward.

We introduce \method, an Explore-then-Integrate approach to self-improvement. \method first generates a portfolio of focused and diverse improvement hypotheses grounded in concrete limitations of the current agent’s runtime source. Each hypothesis is implemented independently from the same parent, producing multiple executable alternatives. \method then inspects all valid alternatives, identifies mechanisms that are complementary, redundant, conflicting, incomplete, or unnecessarily complex, and integrates compatible mechanisms into one coherent descendant. The integrated descendant is accepted only if deterministic structural and operational checks pass. No downstream task outcomes or task-solving trajectories of evolved candidates are used to guide this process. Figure~\ref{fig:explore_synthesize} illustrates the difference between ELITE and existing evaluate-and-select approaches. On the 60-task subset of SWE-bench Pro, \method{} improves the initial agent from 60.0\% to 85.6\%, compared with 86.7\% for DGM and 88.3\% for HGM, while using zero candidate task evaluations and reducing observed evolution time by 8.8$\times$ and 3.2$\times$, respectively, under our evaluated configurations. In cross-benchmark transfer without further adaptation, the evolved agent achieves 73.3\% on \textit{SWE-bench Verified} and 69.3\% on \textit{Polyglot}. Ablations further show that successful task-fitness-free evolution still requires constraints: removing deterministic validity filtering causes performance to collapse. Post-hoc analyses further examine how integration exploits mechanisms across exploratory children and characterize the resulting evolution dynamics. Additional experiments examine an established mini-swe-agent scaffold, direct joint implementation of seven proposals, and evaluation-based child selection in place of mechanism integration.

Our contributions are threefold. \textbf{(1)} We observe that coding agents' self-modifications are inspectable programs, so which changes to carry forward can be decided from implementation-level evidence rather than downstream task outcomes; we formulate this as the problem of self-improvement without candidate task evaluation.
\textbf{(2)} We introduce \method, which replaces candidate-level selection with same-parent executable exploration and compatibility-aware mechanism-level integration, while using deterministic checks only for operational validity. \textbf{(3)} We provide empirical evidence that coding-agent self-improvement without downstream candidate task evaluation is feasible in our setting, and analyze its evolution dynamics, transfer behavior, and key design components.

\section{Related Work}
\label{sec:related_work}

\paragraph{Self-improvement in coding agents.}
Self-improving agents can modify either the underlying model or the agent scaffold, including prompts, memory, tools, workflows, and control logic \citep{gao2025survey,fang2025comprehensivesurveyselfevolvingai,ren2026self}. Software-engineering agents provide infrastructure for interacting with repositories, development tools, and execution environments \citep{qian2024chatdev,yang2024swe,wang2025openhands,hong2024metagpt,xia2025demystifying}. Automated design methods such as GPTSwarm, ADAS, and AFlow optimize prompts, graph structures, or code-represented workflows \citep{zhuge2024language,hu2025automated,zhang2025aflow}, while self-referential approaches such as STOP, Gödel Agent, and SICA modify their own scaffolds or implementations \citep{zelikman2023self,yin2025godel,robeyns2025self}. Coding agents thus provide a concrete setting for recursive self-improvement because their executable implementations can be directly modified and used as the basis for subsequent improvements.

Research on self-improvement has a long history, including self-referential learning and self-modifying policies \citep{schmidhuber1987selfreferential,schmidhuber1997shifting}, as well as the Gödel Machine's formulation of self-rewriting under provable utility improvement \citep{schmidhuber2003godel}. Within this area, evolutionary search offers one approach to program improvement. Relevant historical foundations include early evolutionary computation \citep{barricelli1954esempi,holland1975adaptation}, genetic programming \citep{smith1980learning,645511.657085,dickmanns1987genetische}, and meta-evolution of the improvement procedure itself \citep{schmidhuber1987selfreferential}. Recent LLM-based systems revisit these ideas through executable self-modification. DGM uses benchmark performance to guide archive expansion \citep{zhang2026darwin}, while HGM estimates lineage potential from descendant outcomes \citep{wang2026huxley}. Hyperagents combines task solving and meta-level improvement in one editable program \citep{zhang2026hyperagents}. GEA and MGM share experience or reuse trajectories and outcomes across groups and lineages \citep{weng2026group,liu2026mendelgodelmachinerecursive}; SIFT reduces evaluation cost through LLM-based ranking \citep{fu2026selfimprovementfasttreesearch}, and RQGM co-evolves evaluation criteria \citep{iacob2026redqueengodelmachine}. Despite their different search and evaluation strategies, these systems retain downstream task-derived evidence to guide evolution. \method{} instead studies whether implementation-level inspection and mechanism integration can construct the next descendant without downstream candidate task evaluation.

\paragraph{Implementation-level reasoning for code changes.}
Recent work in software engineering explores implementation-level reasoning without executing candidate code. ICE-Score and CodeJudge assess generated code, while RePaCA reasons over buggy and patched implementations to assess patch correctness \citep{zhuo2024ice,tong2024codejudge,fuster2026repaca}. LLM critics assess repository-level changes, sometimes using reference information such as gold test patches \citep{yadavally2025large}. Agentic Rubrics scores patches against repository-specific criteria \citep{raghavendra2026agentic}, while SWE-RM learns reward models for fine-grained feedback without unit-test execution \citep{shum2026swe}. ASPIRE uses multiple agents to simulate runtime behavior and reason about correctness without code execution \citep{ramanan2024aspire}. PatchFusion combines edit-level evidence across repair candidates without test outcomes at decision time \citep{yang2026single}. These works address feedback, verification, or patch combination for individual tasks, rather than how an agent's own implementation should evolve across generations. \method{} draws on the inspectability of code, using implementation-level reasoning and mechanism integration to construct the next agent without running candidate agents on downstream tasks or using evolved-candidate task outcomes to guide evolution.
\section{Preliminaries}
\label{sec:preliminaries}

\subsection{Formalizing Self-Improving Agents}
\label{sec:formalization}

Following prior work \citep{ren2026self}, we formalize an agent configuration as $\mathcal{A}=(\theta,\Sigma)$, where $\theta$ denotes the parameters of the LLM and $\Sigma$ denotes the agent scaffold, including prompts, memory, tools, and control logic. Self-improvement proceeds over generations $t=0,1,\ldots,T-1$, with updates abstracted as $\mathcal{A}_{t+1}=\operatorname{\mathtt{IMPROVE}}(\mathcal{A}_{0:t};S_t)$, where $S_t$ denotes update-relevant information such as task outcomes, execution trajectories, critiques, verification signals, or modifications \citep{weng2026group,liu2026mendelgodelmachinerecursive,zhang2026outcomefeedbacktrajectoryqualityguidedexperiencebased}. Each updated agent persists as a basis for self-modification. We focus on coding agents with a fixed LLM, so $\mathcal{A}_t=(\theta,\Sigma_t)$ throughout evolution. In our setting, $\Sigma_t$ is implemented as a source-code repository (Appendix~\ref{app:initial_agent}), allowing the scaffold and its modifications to be inspected before downstream evaluation. Such inspection does not establish downstream capability, but can provide implementation-level information \citep{yadavally2025large,raghavendra2026agentic,shum2026swe}.

\subsection{Evaluate-and-Select Self-improvement}
\label{sec:evaluate_select}

Self-improving coding agents organize search differently, but many use downstream task outcomes to guide evolution. We call this pattern \emph{evaluate-and-select}. Let $\mathcal{T}=\{\tau_1,\ldots,\tau_N\}$ be a downstream evaluation set, and let $R(\mathcal{A},\tau_i)$ denote the outcome of agent $\mathcal{A}$ on task $\tau_i$. Under evaluate-and-select, these outcomes, or estimates obtained from partial evaluation, enter $S_t$ and influence which agents, groups, or lineages are retained, expanded, or allocated further evaluation. The selection rule may be deterministic or stochastic, may operate over retained historical agents, and may combine task outcomes with lineage statistics, novelty, or resource costs; it need not select a single highest-scoring winner. This formulation encompasses a range of existing evaluation-driven strategies for selecting or allocating further effort across agents, groups, and lineages \citep{zhang2026darwin,wang2026huxley,weng2026group,liu2026mendelgodelmachinerecursive,fu2026selfimprovementfasttreesearch,iacob2026redqueengodelmachine}.

Obtaining this evidence requires task execution: evaluating $M$ candidates on $N$ tasks costs $MN$ rollouts, while adaptive allocation changes the number of evaluations assigned to each candidate but does not eliminate this cost \citep{fu2026selfimprovementfasttreesearch, dai2026evoallocselfevolvingresourceallocation}. Repeated optimization against a fixed evaluation distribution may further encourage benchmark-specific adaptation \citep{kapoor2024ai}. Separately, agent-level performance scores do not by themselves specify how complementary implementation mechanisms should be combined. These considerations motivate our question: \emph{can coding agents decide which changes to carry forward without downstream candidate task evaluation?}

\section{\method: Explore-then-Integrate Self-Improvement}
\label{sec:method}

Under the formalization in Section~\ref{sec:formalization}, the two paradigms differ in the information used to update the agent. Evaluate-and-select uses downstream task-derived evidence in $S_t$ to guide which agents or lineages receive further evolutionary effort. \method instead uses exploratory implementations and their deterministic validity results; downstream task outcomes of evolved candidates are not used in $S_t$ to guide integration or next-agent decisions. Rather than selecting an entire child based on estimated performance, \method treats implementations as evidence and carries forward compatible mechanisms, shifting the unit of evolutionary decision from whole candidates to implemented mechanisms.

At each generation, \method follows two stages. \textbf{Explore} generates a portfolio of focused improvement hypotheses grounded in limitations of the current agent, each of which is independently implemented from the same parent to produce an executable alternative. \textbf{Integrate} then reasons over the mechanisms realized in these alternatives and reimplements a compatible subset into a coherent descendant. Deterministic gates are used only to ensure structural and operational validity; no downstream benchmark task, task-solving trajectory, or candidate performance score is used to guide the next-agent decision.

\subsection{Limitation-Driven Same-Parent Exploration}
\label{sec:explore}

\paragraph{Limitation-driven hypothesis generation.}
At generation $t$, \method examines the implementation of the current agent $\mathcal{A}_t$ together with a compact evolution history $m_t$, which records prior self-modification instructions and their accompanying code-based rationales. Rather than applying a fixed set of mutation categories, a proposal operator $\mathcal{P}$ jointly generates $K$ focused improvement hypotheses,
\begin{equation}
    \mathcal{H}_t=\mathcal{P}(\mathcal{A}_t,m_t;K)=\{h_t^{(1)},\ldots,h_t^{(K)}\}.
\end{equation}
Each hypothesis specifies the capability to improve, a concrete limitation in the current implementation, a focused mechanism for addressing it, and the conditions under which the implementation is complete (Appendix~\ref{app:hypothesis_generation_prompts}).

\paragraph{Same-parent executable exploration.}
Each hypothesis is implemented independently against the same parent $\mathcal{A}_t$. An implementation operator $\mathcal{E}$ produces a patch $\Delta_t^{(k)}$, which is applied to $\mathcal{A}_t$ to obtain an executable exploratory child:
\begin{equation}
    \Delta_t^{(k)}
    =
    \mathcal{E}(\mathcal{A}_t,h_t^{(k)}),
    \qquad
    \mathcal{A}_t^{(k)}
    =
    \operatorname{Apply}(\mathcal{A}_t,\Delta_t^{(k)}),
    \qquad
    k=1,\ldots,K.
\end{equation}
Because all exploratory children are independently instantiated from the same parent, their modifications provide directly comparable implementation-level alternatives rather than sequentially accumulated changes. This common reference point also makes it possible to reason about mechanisms across children before constructing the next descendant.

\paragraph{Deterministic validity filtering.}
Before integration, \method applies a lightweight deterministic gate $G_{\mathrm{exp}}(\mathcal{A})\in\{0,1\}$ to exclude invalid program-level modifications. The gate checks structural and operational validity, such as executability, deployability, required agent interfaces, and permitted modification scope. Python scripts perform compilation, imports, and predefined assertions; Appendix~\ref{app:gate_implementation} details the checks. The valid alternatives form
\begin{equation}
    \mathcal{V}_t
    =
    \left\{
        \mathcal{A}_t^{(k)}
        \mid
        G_{\mathrm{exp}}(\mathcal{A}_t^{(k)})=1
    \right\}.
\end{equation}
Importantly, $G_{\mathrm{exp}}$ is an admissibility check rather than a measure of downstream capability: it uses no benchmark task outcomes or candidate performance scores.

\paragraph{From exploration to integration.}
Exploration therefore produces a set of independently implemented, operationally valid alternatives rather than a ranked population. Since no downstream task score identifies a winning child in $\mathcal{V}_t$, the remaining decision is not which child should continue as a whole, but which mechanisms implemented across these children can be carried forward together. Section~\ref{sec:integrate} addresses this decision through compatibility-aware mechanism integration.

\subsection{Compatibility-Aware Mechanism Integration}
\label{sec:integrate}

\paragraph{From children to mechanisms.}
Given the valid alternatives $\mathcal{V}_t$, \method shifts the decision unit from whole children to their implemented mechanisms. Rather than assigning each child a scalar score or predicting which child would perform best, the integrator treats the implementations themselves as evidence. Mechanisms are identified from the intended improvement $h_t^{(k)}$ together with its realized implementation, allowing a mechanism to be considered independently of the child that contains it and jointly with compatible mechanisms realized by other children. For each valid child in $\mathcal{V}_t$, the integrator receives its hypothesis $h_t^{(k)}$, a summary of the realized patch $\Delta_t^{(k)}$, and the gate results; the full patch remains available for inspection. The integrator first reviews this evidence across all valid children and then inspects the full implementations of a small set of relevant alternatives. The underlying LLM performs integration through implementation-level reasoning guided by the structured prompt in Appendix~\ref{app:integration_prompt}, without a separate compatibility model or task-performance predictor.

\paragraph{Compatibility-aware mechanism reasoning.}
Using this implementation-level evidence, the integrator reasons about candidate mechanisms along five dimensions: \textit{complementary}, when they can operate together; \textit{redundant}, when they substantially overlap; \textit{conflicting}, when they make incompatible assumptions about interfaces, state, or control flow; \textit{overly complex}, when they introduce unnecessary machinery; and \textit{incomplete}, when the intended mechanism is only partially realized. These judgments determine which mechanisms can be carried forward together, without reducing each child to a single predicted-quality score. Based on these judgments, an integration operator $\mathcal{I}$ constructs an integrated patch against the original parent:
\begin{equation}
    \Delta_t^{\mathrm{int}}
    =
    \mathcal{I}(\mathcal{A}_t,\mathcal{V}_t),
    \qquad
    \mathcal{A}_t^{\mathrm{int}}
    =
    \operatorname{Apply}(\mathcal{A}_t,\Delta_t^{\mathrm{int}}).
\end{equation}
Importantly, integration operates at the mechanism level rather than by mechanically concatenating or sequentially applying exploratory patches. The selected mechanisms are reimplemented as a coherent modification of the common parent $\mathcal{A}_t$, allowing shared interfaces, dependencies, and control flow to be reconciled when mechanisms interact.

\paragraph{Deterministic descendant acceptance.}
The integrated descendant is accepted only if it passes a stricter deterministic gate $G_{\mathrm{acc}}(\mathcal{A})\in\{0,1\}$, which verifies structural and operational validity. It adds smoke tests of the tool-call protocol and main execution flow. Failure retains the current parent; details appear in Appendix~\ref{app:gate_implementation}. The next generation is therefore
\begin{equation}
    \mathcal{A}_{t+1}
    =
    \begin{cases}
        \mathcal{A}_t^{\mathrm{int}},
        & G_{\mathrm{acc}}(\mathcal{A}_t^{\mathrm{int}})=1,\\
        \mathcal{A}_t,
        & \text{otherwise}.
    \end{cases}
\end{equation}
As with $G_{\mathrm{exp}}$, this gate verifies structural and operational validity rather than downstream capability. No downstream benchmark task, task-solving trajectory, or candidate performance score is used to determine which mechanisms are integrated or whether the resulting descendant is accepted as the next agent.

\paragraph{Downstream evaluation cost.}
Unlike the direct evaluate-and-select setting in Section~\ref{sec:evaluate_select}, \method requires no downstream candidate-task rollouts for next-agent decisions. Its cost instead comes from generating executable alternatives, inspecting their implementations, and constructing the integrated descendant; this decision cost is independent of the size of a downstream evaluation set. The complete procedure is summarized in Algorithm~\ref{alg:elite} in Appendix~\ref{app:elite_algorithm}.
\section{Experiments}
\label{sec:experiments}

\subsection{Experimental Setup}
\label{sec:setup}

\paragraph{Benchmarks.}
We evaluate on \textit{SWE-bench Pro}, \textit{SWE-bench Verified} \citep{deng2025swe, jimenez2024swe}, and \textit{Polyglot} \citep{gauthier2024polyglot}. We use a fixed 60-task subset of SWE-bench Pro for the evolution comparison, while SWE-bench Verified and Polyglot serve as transfer benchmarks. After evolution, the resulting agents are frozen and evaluated on the same Pro subset and the two transfer subsets without further adaptation. Detailed benchmark and task-subset information is provided in Appendix~\ref{app:benchmark_details}.

\paragraph{Compared methods and evolution protocol.}
We compare \method with DGM \citep{zhang2026darwin} and HGM \citep{wang2026huxley} using \texttt{DeepSeek-V4-Flash} and \texttt{Qwen-3.7-Plus}. DGM and HGM use outcomes from the Pro subset as evolutionary feedback, whereas \method uses no downstream task outcomes during evolution. For each backbone, all methods start from the same scaffold and use the same LLM for both self-modification and task solving, holding the starting agent and model fixed across methods. All three methods have three independent DeepSeek evolution runs; Qwen has three \method runs and one run per baseline. A full evolution run is the unit of replication. Run-level results and sample SDs are in Appendix~\ref{app:elite_repeated_runs}; configurations are in Appendix~\ref{app:experimental_settings}.

\paragraph{Metrics.}
We report final task success rate as the primary performance metric. To characterize the resources consumed during self-improvement, we additionally report downstream task-evaluation count, self-modification throughput (Self-Mod./h), estimated evolution-stage token usage, end-to-end wall-clock time, and estimated API cost.

\subsection{Main Results}
\label{sec:main_results}

\paragraph{Performance.}
Table~\ref{tab:main_results} summarizes performance across both LLM backbones. Under \texttt{DeepSeek-V4-Flash}, \method{} improves Pro performance from 60.0\% to 85.6\%, slightly below DGM and HGM at 86.7\% and 88.3\%, respectively, which use downstream task evaluation feedback to guide evolution. However, its advantage emerges when directly evaluated on \textit{SWE-bench Verified} and \textit{Polyglot} without further evolution, achieving 73.3\% and 69.3\%, respectively. These results indicate that the improvements produced by \method{} transfer across benchmarks despite the absence of task-specific evaluation feedback during evolution. In contrast, DGM and HGM, evolved on Pro, show limited gains or even regression on these transfer benchmarks. The same pattern holds under \texttt{Qwen-3.7-Plus}, although the trade-off becomes more pronounced: evolving without downstream task feedback is associated with a larger performance gap behind DGM and HGM on Pro (45.6\% vs.\ 50.0\% and 48.3\%), while \method{} still achieves better cross-benchmark transfer (56.7\% on Verified and 61.3\% on Polyglot).

\begin{table}[htbp]
\centering
\normalsize
\setlength{\tabcolsep}{4pt}
\renewcommand{\arraystretch}{1.12}
\caption{\textbf{Main performance results across LLM backbones.}
Scores are success rates (\%); superscripts denote changes from the corresponding initial agent in percentage points. Shaded Avg. columns report mean scores across the three benchmarks. DeepSeek results are three-run means; Qwen results are three-run means for \method{} and single runs for DGM/HGM.}
\label{tab:main_results}

\begin{tabular*}{\linewidth}{
@{\extracolsep{\fill}}
lccc
>{\columncolor{avggray}}c
ccc
>{\columncolor{avggray}}c
@{}}
\toprule
& \multicolumn{4}{c}{\texttt{DeepSeek-V4-Flash}}
& \multicolumn{4}{c}{\texttt{Qwen-3.7-Plus}} \\
\cmidrule(lr){2-5}\cmidrule(lr){6-9}
\textbf{Method}
& \textbf{Pro} & \textbf{Verified} & \textbf{Polyglot} & \textbf{Avg.}
& \textbf{Pro} & \textbf{Verified} & \textbf{Polyglot} & \textbf{Avg.} \\
\midrule
\textit{Initial Agent}
& 60.0 & 64.0 & 64.0 & 62.7
& 33.3 & 48.0 & 50.0 & 43.8 \\
\midrule
DGM
& 86.7\gain{26.7}
& 63.3\loss{0.7}
& 62.7\loss{1.3}
& 70.9\gain{8.2}
& \textbf{50.0}\gain{16.7}
& 54.0\gain{6.0}
& 50.0\nogain
& 51.3\gain{7.6} \\

HGM
& \textbf{88.3}\gain{28.3}
& 65.3\gain{1.3}
& 64.7\gain{0.7}
& 72.8\gain{10.1}
& 48.3\gain{15.0}
& 54.0\gain{6.0}
& 58.0\gain{8.0}
& 53.4\gain{9.7} \\

\textbf{\method}
& 85.6\gain{25.6}
& \textbf{73.3}\gain{9.3}
& \textbf{69.3}\gain{5.3}
& \textbf{76.1}\gain{13.4}
& 45.6\gain{12.3}
& \textbf{56.7}\gain{8.7}
& \textbf{61.3}\gain{11.3}
& \textbf{54.5}\gain{10.7} \\
\bottomrule
\end{tabular*}
\end{table}

\begin{wraptable}{R}{0.55\textwidth}
\vspace{-2.0\baselineskip}
\centering
\small
\setlength{\tabcolsep}{2pt}
\renewcommand{\arraystretch}{1.10}
\caption{\textbf{Evolution resource usage across LLM backbones.}
Times are observed wall-clock durations under the reported execution settings. Baseline resource measurements are from the original runs.}
\label{tab:evolution_cost}
\begin{tabular*}{\linewidth}{@{\extracolsep{\fill}}lccccc@{}}
\toprule
\textbf{Method}
& \shortstack{\textbf{Task}\\\textbf{Eval.} $\downarrow$}
& \shortstack{\textbf{Self-Mod.}\\\textbf{/h} $\uparrow$}
& \shortstack{\textbf{Est. Tokens}\\\textbf{(M)} $\downarrow$}
& \shortstack{\textbf{Time}\\\textbf{(h)} $\downarrow$}
& \shortstack{\textbf{Cost}\\\textbf{(\$)} $\downarrow$} \\
\midrule
\rowcolor{ModelGray}
\multicolumn{6}{c}{\texttt{DeepSeek-V4-Flash}} \\
DGM & 810 & 0.58 & 2900 & 43.15 & 45 \\
HGM & 800 & 3.77 & 2100 & 15.66 & 30 \\
\textbf{\method}
& \textbf{0} & \textbf{16.35} & \textbf{80}
& \textbf{4.90} & \textbf{8} \\
\midrule
\rowcolor{ModelGray}
\multicolumn{6}{c}{\texttt{Qwen-3.7-Plus}} \\
DGM & 820 & 0.58 & 2500 & 48.33 & 125 \\
HGM & 800 & 2.99 & 1800 & 19.70 & 90 \\
\textbf{\method}
& \textbf{0} & \textbf{10.89} & \textbf{70}
& \textbf{7.35} & \textbf{20} \\
\bottomrule
\end{tabular*}
\vspace{-0.8\baselineskip}
\end{wraptable}

\paragraph{Evolution efficiency.}
Table~\ref{tab:evolution_cost} reports evolution resource usage. Under \texttt{DeepSeek-V4-Flash}, \method{} completes evolution in a mean of 4.90 hours, compared with 43.15 hours for DGM and 15.66 hours for HGM, yielding observed wall-clock speedups of 8.81$\times$ and 3.20$\times$ under the reported execution settings. A key source of this efficiency is the elimination of candidate task evaluations, compared with 810 for DGM and 800 for HGM. These task rollouts consume additional time, tokens, and API expenditure during evolution. \method{} achieves 16.35 self-modification attempts per hour, compared with 0.58 for DGM and 3.77 for HGM. Under \texttt{Qwen-3.7-Plus}, \method{} similarly avoids candidate task evaluations, averaging 7.35 hours, approximately \$20, and 10.89 self-modification attempts per hour, with lower estimated token usage than both baselines.

\subsection{Transfer and Evolution Analysis}
\label{sec:generalization}
\label{sec:posthoc_quality}
\label{sec:evolution_trace}

\begin{figure}[htbp]
    \centering
    \includegraphics[width=1.0\linewidth]{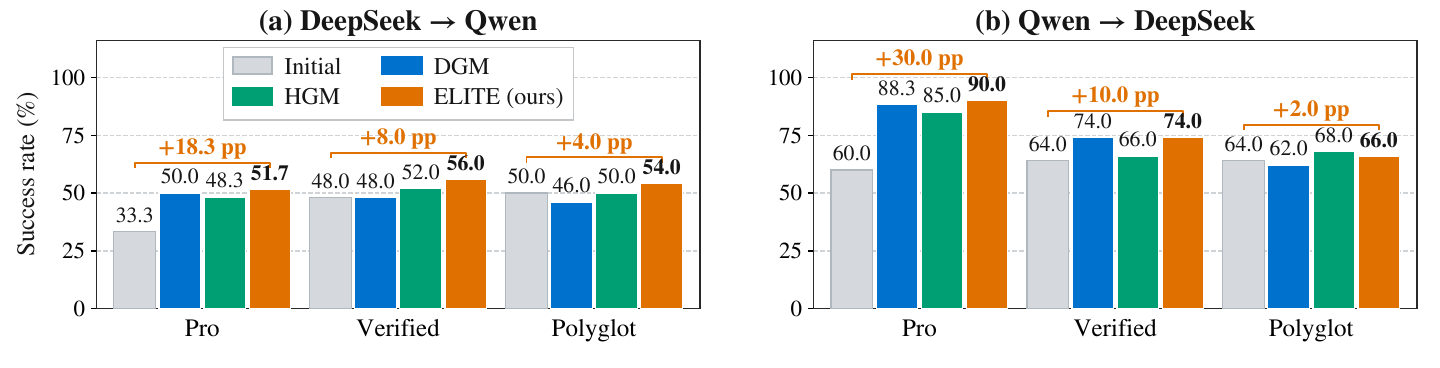}
    \caption{\textbf{Cross-backbone scaffold portability.} Scaffolds evolved in the SWE-bench Pro setting by DGM, HGM, and \method{} with one LLM backbone are transferred to the other backbone without further evolution and evaluated on Pro, Verified, and Polyglot.}
    \label{fig:scaffold_portability}
\end{figure}

\paragraph{Cross-backbone scaffold portability.}
We further evaluate cross-backbone scaffold portability by pairing scaffolds evolved with one backbone with the other backbone, without additional evolution. All scaffolds are evolved only in the SWE-bench Pro setting; SWE-bench Verified and Polyglot are used solely for transfer evaluation and never participate in evolution. As shown in Figure~\ref{fig:scaffold_portability}, the DeepSeek-evolved \method{} scaffold achieves the best transferred performance on all three benchmarks under \texttt{Qwen-3.7-Plus}. In the reverse direction, the Qwen-evolved \method{} scaffold likewise achieves the best or second-best transferred performance across the three benchmarks. Overall, these results provide evidence that some scaffold improvements produced by \method{} remain useful when transferred across the two evaluated backbones.

\begin{figure}[!htbp]
    \centering
    \includegraphics[width=\linewidth]{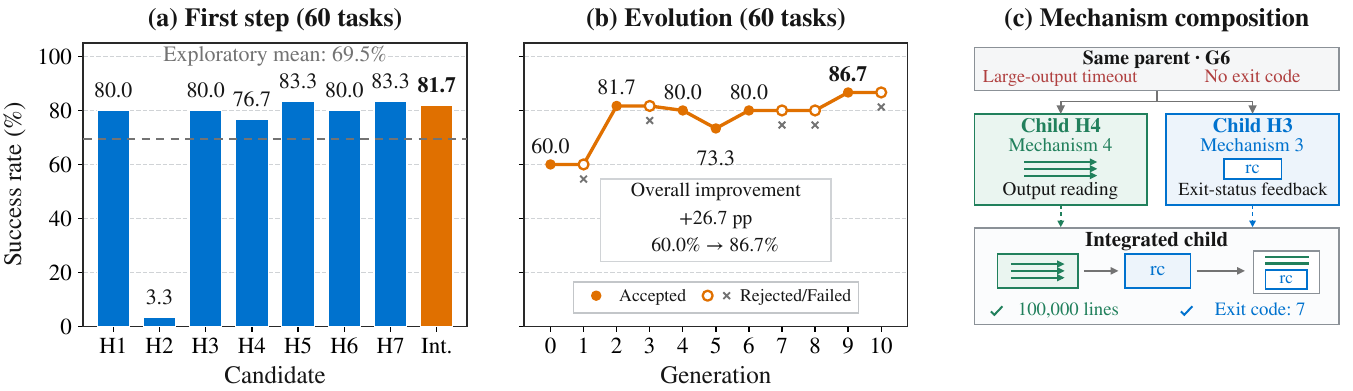}
    \caption{\textbf{Post-hoc evolution dynamics and mechanism composition in a \texttt{DeepSeek-V4-Flash} run.}
    (a) First-step scores on 60 Pro tasks for seven exploratory children and the integrated descendant; the dashed line marks the child mean.
    (b) Retained-agent performance over ten generations, starting from the initial agent at generation 0. Filled markers denote accepted updates. Crosses mark generations that did not update the parent: hypothesis composition failed before integration at generations 1, 3, and 7, whereas integrated descendants were rejected by the acceptance gate at generations 8 and 10.
    (c) A mechanism composition case: integration combines H4's output reading with H3's exit-status feedback, returning 100,000 lines and exit code 7 in a local probe.}
    \label{fig:posthoc_outcomes}
\end{figure}

\paragraph{Post-hoc evolution dynamics.}
We analyze the evolution process after it has concluded; all downstream task outcomes reported in this analysis are measured post hoc and never exposed to \method{} during evolution. Figure~\ref{fig:posthoc_outcomes}(a) compares the seven exploratory children and the integrated descendant from the first accepted step on the same 60 SWE-bench Pro tasks. The exploratory children exhibit substantial outcome variation: six achieve 76.7--83.3\%, while one falls to 3.3\%. The integrated descendant reaches 81.7\%, outperforming five of the seven children while remaining below the highest exploratory score of 83.3\%. This comparison illustrates that \method{} can construct a competitive descendant without observing candidate task performance, although integration does not necessarily recover the best-performing exploratory alternative.

Figure~\ref{fig:posthoc_outcomes}(b) shows the complete $T=10$ evolution process. The retained agent is updated only when an integrated descendant passes the acceptance gate; otherwise, the current agent is carried forward to the next generation. Across five accepted updates, post-hoc task performance improves overall from 60.0\% to 86.7\%, despite two intermediate regressions. After the first accepted update reaches 81.7\%, subsequent evolution adds a further 5.0 percentage points. This non-monotonic trajectory is consistent with the gate's intended role: it enforces structural and operational validity without guaranteeing downstream task improvement. An additional sampling-based comparison between integrated descendants and exploratory populations is provided in Appendix~\ref{app:posthoc_sampling}.

\paragraph{Mechanism-level integration.}
We further inspect the implementations to examine what integration carries forward. Figure~\ref{fig:posthoc_outcomes}(c) presents a representative composition case in which two same-parent exploratory children independently introduce complementary mechanisms: one enables complete output reading, while the other incorporates exit-status feedback. Rather than selecting either child as the next agent, \method{} integrates both mechanisms into a single coherent descendant. In the analyzed DeepSeek run, integration incorporates primary mechanisms from 25 of 41 eligible proposals. Together, the case study and incorporation analysis illustrate how mechanisms from different exploratory children can be carried forward into one descendant without ranking children by downstream task performance. Additional composition and refinement cases are provided in Appendix~\ref{app:combined_cases}; proposal diversity, gate outcomes, and repository-level analyses are reported in Appendix~\ref{app:evolution_trace}.

\paragraph{Effectiveness across initial scaffolds.}
\label{sec:mini_swe_results}

\begin{wraptable}{r}{0.55\textwidth}
\vspace{-1.7\baselineskip}
\centering
\footnotesize
\setlength{\tabcolsep}{3pt}
\renewcommand{\arraystretch}{1.15}
\caption{\textbf{ELITE evolution from mini-swe-agent.}
Scores are success rates (\%); gains are in percentage points.
Child indices follow acceptance order; child 6 is the final agent
after ten evolution rounds. Dashes indicate unreported scores.}
\label{tab:mini_swe_agent}

{
\begin{tabularx}{\linewidth}{
    @{}
    >{\raggedright\arraybackslash}X
    ccc
    @{}
}
\toprule
\textbf{Agent}
& \textbf{Pro}
& \textbf{Verified}
& \textbf{Polyglot} \\
\midrule
\textit{Initial agent}
& 71.7 & 70.0 & 72.0 \\
Accepted child 1
& 73.3 & --- & --- \\
Accepted child 3
& 80.0 & --- & --- \\
Accepted child 6 (final)
& \textbf{83.3} & \textbf{78.0} & \textbf{78.0} \\
\midrule
Initial-to-final gain
& +11.7 & +8.0 & +6.0 \\
\bottomrule
\end{tabularx}
}

\vspace{-0.5\baselineskip}
\end{wraptable}

To examine ELITE's effectiveness across initial scaffolds, we additionally evaluate it on mini-swe-agent\footnote{We use v2.4.6, the latest release at the time of our experiments: \url{https://github.com/SWE-agent/mini-swe-agent/releases/tag/v2.4.6}.} with a fixed \texttt{Claude-opus-5.5} backbone. Ten evolution rounds yield six accepted children. Table~\ref{tab:mini_swe_agent} reports scores for the initial agent and selected accepted children. Pro improves from 71.7\% initially to 73.3\%, 80.0\%, and 83.3\% for accepted children 1, 3, and 6, respectively. The final agent also improves Verified from 70.0\% to 78.0\% and Polyglot from 72.0\% to 78.0\%. These results extend ELITE's effectiveness beyond the DGM starting scaffold, showing further improvement of an established agent over multiple accepted updates and gains on all three benchmarks.

\subsection{Ablation and Sensitivity Analysis}
\label{sec:ablations}

\begin{table}[!htbp]
\centering
\color{black}
\small
\setlength{\tabcolsep}{4pt}
\renewcommand{\arraystretch}{1.06}
\caption{\textbf{Design ablations of \method.}
All methods use DeepSeek. Scores are success rates (\%); Avg. equally weights the three benchmarks. Full \method and the no-gates ablation report three-run means; all other variants use one run. Bold numbers indicate column maxima.}
\label{tab:ablations}

\begin{tabular*}{\linewidth}{@{\extracolsep{\fill}}lcccc@{}}
\toprule
\textbf{Method}
& \textbf{Pro}
& \textbf{Verified}
& \textbf{Polyglot}
& \textbf{Avg.} \\
\midrule
\textbf{Full \method ($K=7$)}
& 85.6 & \textbf{73.3} & \textbf{69.3} & \textbf{76.1} \\
\midrule

\rowcolor{avggray}
\multicolumn{5}{l}{\emph{Alternative selection strategies}} \\
Select-one (LLM-based)
& 83.3 & 68.0 & 64.0 & 71.8 \\
SIFT-style Select-one
& 85.0 & 72.0 & 46.0 & 67.7 \\
Evaluation-based Select-one
& \textbf{91.7} & 70.0 & 62.0 & 74.6 \\
\midrule

\rowcolor{avggray}
\multicolumn{5}{l}{\emph{Exploration and integration ablations}} \\
Single exploration ($K=1$)
& 81.7 & 64.0 & 64.0 & 69.9 \\
Direct joint implementation
& 78.3 & 72.0 & 66.0 & 72.1 \\
\midrule

\rowcolor{avggray}
\multicolumn{5}{l}{\emph{Other component ablations}} \\
w/o hypothesis generation
& 85.0 & 68.0 & 60.0 & 71.0 \\
w/o deterministic gates
& 59.4 & 51.3 & 45.3 & 52.0 \\
\bottomrule
\end{tabular*}
\end{table}

\paragraph{Design ablations.}
Table~\ref{tab:ablations} compares \method{} with seven variants covering child selection, exploration and integration, and other components.
\emph{Select-one} retains $K=7$ same-parent exploration but replaces integration with an LLM decision selecting one child.
\emph{SIFT-style Select-one} instead selects a child using pairwise LLM judgments aggregated with a Bradley--Terry model \citep{fu2026selfimprovementfasttreesearch}.
\emph{Evaluation-based Select-one} evaluates the seven independently implemented children on the same 30 Pro tasks within each round and selects the highest-scoring child instead of integrating them.
\emph{Single exploration} sets $K=1$ and directly uses the exploratory child as the next agent, removing both multi-branch exploration and integration.
\emph{Direct joint implementation} generates seven improvement proposals and implements them jointly as one child, omitting the separate exploratory implementations.
\emph{w/o hypothesis generation} removes structured hypothesis generation and directly proposes self-modifications, retaining $K=7$, integration, and both gates.
\emph{w/o deterministic gates} removes both validity gates, allowing all exploratory children into integration and accepting the resulting descendant without deterministic filtering.
Further implementation details and budgets are provided in Appendix~\ref{app:additional_revision_settings}.

\begin{wraptable}{r}{0.48\textwidth}
\vspace{-0.8\baselineskip}
\centering
\footnotesize
\setlength{\tabcolsep}{3pt}
\renewcommand{\arraystretch}{1.15}
\caption{\textbf{Sensitivity to exploration width.} We use $K=7$ by default.}
\label{tab:exploration_width}
\begin{tabularx}{\linewidth}{
@{}
l
*{4}{>{\centering\arraybackslash}X}
@{}}
\toprule
\textbf{Width $K$}
& \textbf{3}
& \textbf{5}
& \textbf{7}
& \textbf{9} \\
\midrule
Pro      & 82.8 & 85.0 & 85.6 & 88.9 \\
Verified & 67.3 & 70.7 & 73.3 & 70.0 \\
Polyglot & 61.3 & 68.7 & 69.3 & 68.0 \\
\midrule
\rowcolor{avggray}
\textbf{Avg.}
& 70.5 & 74.8 & \textbf{76.1} & 75.6 \\
\bottomrule
\end{tabularx}
\vspace{-0.8\baselineskip}
\end{wraptable}

\paragraph{Integration versus child selection.}
With seven independently implemented children retained, \method averages 76.1\%, exceeding Select-one (71.8\%) and SIFT-style Select-one (67.7\%). Evaluation-based Select-one achieves higher target Pro performance (91.7\% versus 85.6\%), while \method has higher cross-benchmark transfer scores on Verified (73.3\% versus 70.0\%) and Polyglot (69.3\% versus 62.0\%). These observed results support mechanism integration as a competitive alternative to the evaluated selection strategies without candidate task feedback; the selection comparisons are limited to single runs.

\paragraph{The role of executable exploration.}
Direct joint implementation generates seven proposals but implements them jointly without independent executable children, reducing the average from 76.1\% to 72.1\%. This supports obtaining concrete implementation-level evidence before integration. Single exploration scores 69.9\%, reflecting both reduced exploration breadth and removal of multi-child integration; it does not match full \method's candidate count or total compute.

\paragraph{Other components.}
Removing structured hypothesis generation reduces the average to 71.0\%, while removing deterministic gates reduces it to 52.0\%. These results support their complementary roles.

\paragraph{Sensitivity to exploration width.}
We vary the number of same-parent exploratory children $K\in\{3,5,7,9\}$ while keeping the remaining settings fixed. Table~\ref{tab:exploration_width} reports three-run means, with $K=7$ as the default configuration. Among the tested widths, $K=9$ achieves the highest Pro score of 88.9\%, while $K=7$ performs best on both transfer benchmarks. Overall, the average score increases from 70.5\% at $K=3$ to 76.1\% at $K=7$ and remains comparable at 75.6\% for $K=9$, suggesting diminishing returns from further increasing exploration width rather than a monotonic benefit from generating more exploratory children.
\section{Conclusion}
\label{sec:conclusion}

We introduced \method, an Explore-then-Integrate framework for studying whether coding-agent self-improvement can proceed without downstream candidate task evaluation. Rather than using task outcomes to rank evolved candidates, \method explores multiple executable modifications from the same parent and constructs the next agent by reasoning over their implementation mechanisms. Our results provide evidence for the feasibility of substantial coding-agent self-improvement without downstream candidate task feedback in the setting studied here.

\paragraph{Limitations.}
Our evaluation focuses on coding agents and selected benchmark subsets, with limited repetitions for some comparisons. Broader task coverage and additional runs would help characterize variability across settings. \method{} relies on implementation-level reasoning
and validity checks to guide evolution; downstream performance can fluctuate across accepted updates, making longer-term evolution an important subject for further study. The current implementation also maintains a single lineage. Extending it with an archive of historical
agents could preserve alternative evolutionary paths and support a broader investigation of search diversity. Appendix~\ref{app:mechanism_and_scope} provides further discussion.

\bibliographystyle{plainnat}
\bibliography{references}

\clearpage
\appendix
\startcontents[appendices]
\section*{Appendix Contents}
\printcontents[appendices]{}{2}{}
\clearpage

\begingroup
  \let\clearpage\relax
  \section{Appendix}
\label{app:additional_experiments}

\subsection{Additional Results}
\label{app:additional_results}

\subsubsection{Detailed Evolution Traces}
\label{app:evolution_trace}

\paragraph{Repository-level performance.}
Figure~\ref{fig:repository_dynamics} shows that \method's aggregate
gains are distributed across multiple software repositories.
Relative to the corresponding initial agent, the final DeepSeek
scaffold improves on 8 of 11 repositories, ties on 2, and declines
on 1; the Qwen scaffold improves on 7, ties on 2, and declines on 2.
These results demonstrate the breadth of the observed improvements,
despite non-monotonic intermediate DeepSeek scores.
To limit evaluation cost, Qwen post-hoc evaluation covers only
the initial and final agents; intermediate checkpoints are omitted.

\begin{figure}[htbp]
    \centering
    \includegraphics[width=1.0\linewidth]{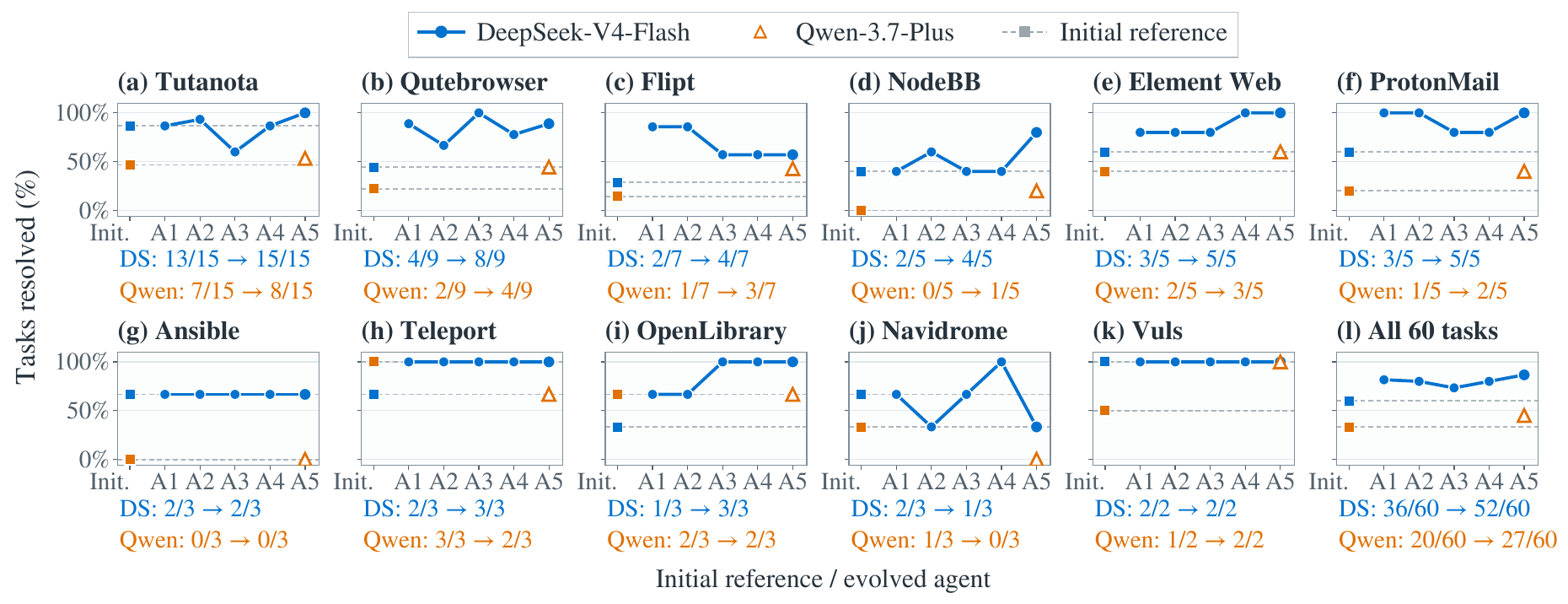}
    \caption{\textbf{Repository-level performance across evolution.}
Post-hoc SWE-bench Pro solve rates across 11 repositories and the aggregate 60-task set. Blue circles denote DeepSeek-V4-Flash checkpoints; orange triangles denote the final Qwen-3.7-Plus checkpoint. Colored squares and gray dashed lines mark initial references. Annotations report initial-to-final solved counts.}
    \label{fig:repository_dynamics}
\end{figure}

\paragraph{Integration selectivity and gate outcomes.}
Figure~\ref{fig:integration_selection}(a) summarizes primary-mechanism incorporation based on manual inspection of the recorded patches and integration logs. Filled bars denote eligible proposals whose primary mechanism is incorporated into the integrated descendant, while hatched bars denote proposals not incorporated. Figure~\ref{fig:integration_selection}(b) reports deterministic gate outcomes: blue denotes passed candidates and hatched orange denotes rejected candidates. The exploration gate accepts 41/48 candidates and the acceptance gate accepts 5/7 integrated descendants. The symbol $\dagger$ marks integration attempts whose resulting descendant is rejected by the acceptance gate. The recorded exploration rejections consist of six conservative type-check false positives and one file-scope violation; the two acceptance rejections arise from a tool-order violation and a call-limit violation.

\begin{figure}[htbp]
    \centering
    \includegraphics[width=0.75\linewidth]{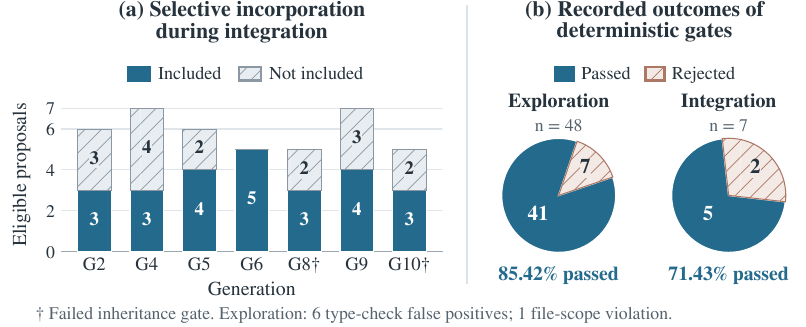}
    \caption{\textbf{Integration choices and deterministic gate outcomes in the \texttt{DeepSeek-V4-Flash} run.} Left: incorporation of primary mechanisms from eligible exploratory proposals. Right: pass and rejection outcomes of the exploration and integration gates.}
    \label{fig:integration_selection}
\end{figure}

\paragraph{Exploration diversity.}
Figure~\ref{fig:diverse} visualizes all seven integration attempts in the \texttt{DeepSeek-V4-Flash} run. Each row corresponds to one shared parent; H1--H7 are independently implemented exploratory children, and the rightmost card is the resulting integrated descendant. The card title summarizes the primary mechanism of each patch. Colored squares indicate which scaffold components are modified, following the legend for Agent, Protocol, Bash, Editor, Prompt, and Utilities. A red border with $\times$ marks an exploratory child rejected by the deterministic gate. The annotation $L$ denotes the total number of added and deleted lines, while $h$ denotes the number of diff hunks. The black bar visualizes $\log(1+L)$ so that patch sizes with different scales remain visually comparable. For each generation, the reported mean diff distance summarizes textual dissimilarity among exploratory patches, shown for all returned children and for gate-passing children separately. The blue strip beneath each integrated descendant shows its textual similarity to H1--H7, ranging from 0 to 1. These quantities describe implementation structure only and should not be interpreted as semantic quality or causal contribution.

\begin{figure}[htbp]
    \centering
    \includegraphics[width=0.9\linewidth]{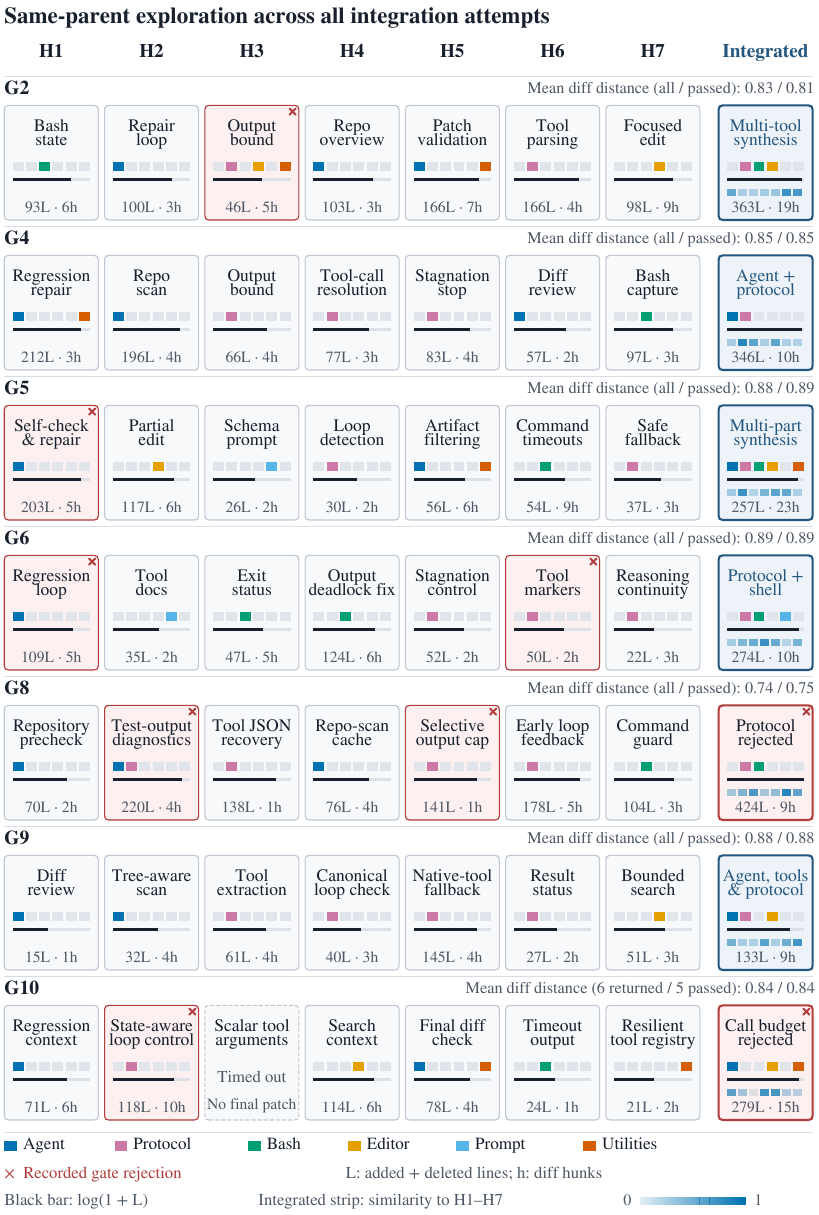}
    \caption{\textbf{Same-parent exploration across integration attempts.} Each row shows H1--H7 independently implemented from a shared parent and the resulting integrated descendant. Card annotations summarize mechanisms and modified components; red crosses mark gate rejections. $L$ denotes added/deleted lines, $h$ diff hunks, black bars show $\log(1+L)$, blue strips show textual similarity to H1--H7, and row labels report mean diff distances.}
    \label{fig:diverse}
\end{figure}

\subsubsection{Results Across Independent Runs}
\label{app:elite_repeated_runs}

Table~\ref{tab:elite_run_performance} reports the individual performance results underlying the three-run \method{} means in the main text. For each backbone, Run~1--3 denote the three reported runs. We report arithmetic means $\pm$ sample standard deviations (SD, $n=3$, with an $n-1$ denominator). Table~\ref{tab:baseline_repeated_runs} adds the DeepSeek baseline repetitions.

\begin{table}[htbp]
\centering
\normalsize
\setlength{\tabcolsep}{4pt}
\renewcommand{\arraystretch}{1.12}
\caption{\textbf{Per-run performance of \method.}
Scores are success rates (\%); SDs are in percentage points.
Verified and Polyglot are evaluated directly after evolution without further adaptation.}
\label{tab:elite_run_performance}
\begin{tabular*}{\linewidth}{@{\extracolsep{\fill}}lcccccc@{}}
\toprule
& \multicolumn{3}{c}{\texttt{DeepSeek-V4-Flash}}
& \multicolumn{3}{c}{\texttt{Qwen-3.7-Plus}} \\
\cmidrule(lr){2-4}\cmidrule(lr){5-7}
\textbf{Run}
& \textbf{Pro} & \textbf{Verified} & \textbf{Polyglot}
& \textbf{Pro} & \textbf{Verified} & \textbf{Polyglot} \\
\midrule
Run 1 & 86.7 & 72.0 & 68.0 & 45.0 & 56.0 & 60.0 \\
Run 2 & 85.0 & 74.0 & 68.0 & 45.0 & 58.0 & 62.0 \\
Run 3 & 85.0 & 74.0 & 72.0 & 46.7 & 56.0 & 62.0 \\
\midrule
\textbf{Mean $\pm$ SD}
& $85.6 \pm 1.0$ & $73.3 \pm 1.2$ & $69.3 \pm 2.3$
& $45.6 \pm 1.0$ & $56.7 \pm 1.2$ & $61.3 \pm 1.2$ \\
\bottomrule
\end{tabular*}
\end{table}

\begin{table}[htbp]
\centering
\small
\setlength{\tabcolsep}{5pt}
\caption{DeepSeek baseline repetitions and comparison with \method. Success rates are percentages; sample SDs are percentage points. Run~1 reports the original baseline results.}
\label{tab:baseline_repeated_runs}
{
\begin{tabular*}{\linewidth}{@{\extracolsep{\fill}}lccc>{\columncolor{avggray}}c@{}}
\toprule
Method and run & Pro & Verified & Polyglot & Avg. \\
\midrule
DGM Run 1 & 86.7 & 64.0 & 62.0 & 70.9 \\
DGM Run 2 & 85.0 & 64.0 & 62.0 & 70.3 \\
DGM Run 3 & 88.3 & 62.0 & 64.0 & 71.4 \\
\textbf{DGM Mean $\pm$ SD} & $86.7\pm1.7$ & $63.3\pm1.2$ & $62.7\pm1.2$ & $70.9\pm0.6$ \\
\midrule
HGM Run 1 & 86.7 & 68.0 & 64.0 & 72.9 \\
HGM Run 2 & 88.3 & 62.0 & 64.0 & 71.4 \\
HGM Run 3 & 90.0 & 66.0 & 66.0 & 74.0 \\
\textbf{HGM Mean $\pm$ SD} & $88.3\pm1.7$ & $65.3\pm3.1$ & $64.7\pm1.2$ & $72.8\pm1.3$ \\
\midrule
\textbf{\method Mean $\pm$ SD} & $85.6\pm1.0$ & $73.3\pm1.2$ & $69.3\pm2.3$ & $76.1\pm0.8$ \\
\bottomrule
\end{tabular*}}
\end{table}

\subsubsection{Sampled Exploration--Integration Comparison}
\label{app:posthoc_sampling}

We use a post-hoc diagnostic on one \texttt{DeepSeek-V4-Flash} evolution run to compare integrated candidates with their valid exploratory populations. Evaluations occur after acceptance decisions and never inform evolution. Figure~\ref{fig:posthoc_diagnostic} shows the five accepted updates.

\paragraph{Tasks and sampling.}
We randomly draw 10 SWE-bench Pro tasks independently of agent outcomes and keep this subset fixed across all comparisons. Each integrated candidate is evaluated once per task. Let $m_t\geq2$ denote the number of exploratory children passing the exploration gate at step $t$. Using a fixed random seed, we randomly order tasks and children, then assign two distinct children to each task using a balanced cyclic schedule. The resulting 20 distinct child--task pairs cover all valid children in this study, with evaluation counts differing by at most one between children. Assignments use neither downstream outcomes nor integration decisions.

\begin{figure}[htbp]
    \centering
    \includegraphics[width=.78\linewidth]{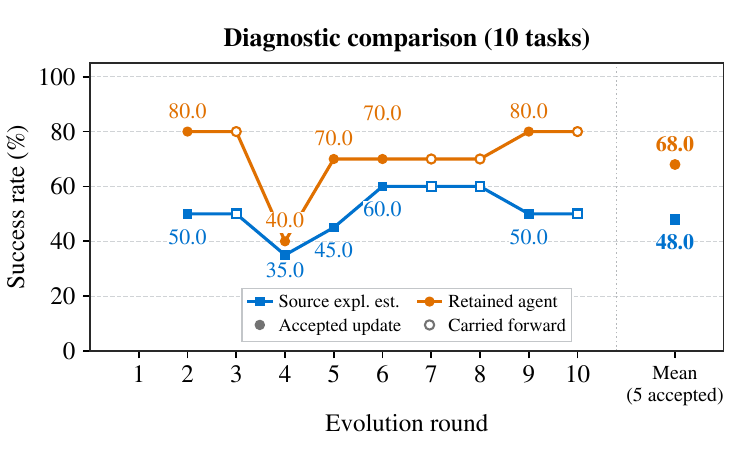}
    \caption{\textbf{Sampled exploration--integration comparison.}
    Filled markers show recorded comparisons at accepted rounds 2, 4, 5, 6, and 9. Hollow markers carry forward the preceding retained agent and its source exploratory estimate after rejection or failure; they are not outcomes of the unaccepted children or newly generated
    exploratory populations. The Mean column equally averages the five accepted comparisons, counting each once.}
    \label{fig:posthoc_diagnostic}
\end{figure}

\paragraph{Reported quantities and evaluation budget.}
The exploration estimate is the mean binary success rate over the 20 sampled child--task outcomes; the integration score is the success rate on the 10 integrated-agent rollouts. Their difference, in percentage points, is the integration gap. Each comparison therefore uses $20+10=30$ rollouts. The complete diagnostic covers seven integration attempts, requiring $7\times30=210$ fresh rollouts; the five accepted comparisons shown in Figure~\ref{fig:posthoc_diagnostic} account for $5\times30=150$ of these.

\paragraph{Results and interpretation.}
The integrated descendant exceeds the sampled exploratory mean at each of the five accepted updates, with step-averaged scores of 68.0\% and 48.0\%, respectively (a 20.0 pp gap). We average these five comparisons equally. Averaging all ten displayed rounds would instead weight checkpoints by their
retention duration. These are descriptive comparisons on one fixed subset and one accepted lineage; they do not establish a general advantage across tasks or runs. Because each child solves only a subset of the diagnostic tasks, the samples estimate population-average performance on this subset and cannot identify an oracle-best child.

\subsubsection{Code and Outcome Case Studies}
\label{app:combined_cases}

The first three cases analyze the \texttt{DeepSeek-V4-Flash} run; the final three analyze the \texttt{Qwen-3.7-Plus} run. $G$ denotes the original generation within each run.

\paragraph{Combining complementary code changes.}
Figure~\ref{fig:g5_combined_case} links two G6 exploratory modifications to the integrated implementation and their observed behavior. H3 adds exit-status feedback, while H4 introduces continuous output reading. Under identical local
probes, H3 reports silent failures but times out on large output; H4 returns all output but omits the exit status. The integrated tool preserves both properties, returning all 100,000 lines together with the correct failure code. These checks replay the archived \texttt{tools/bash.py} implementations with \texttt{(exit 7)} and \texttt{seq 1 100000; (exit 7)}, each under a 3-second timeout, without LLM calls or benchmark task evaluation.

\begin{figure}[htbp]
    \centering
    \includegraphics[width=0.9\linewidth]{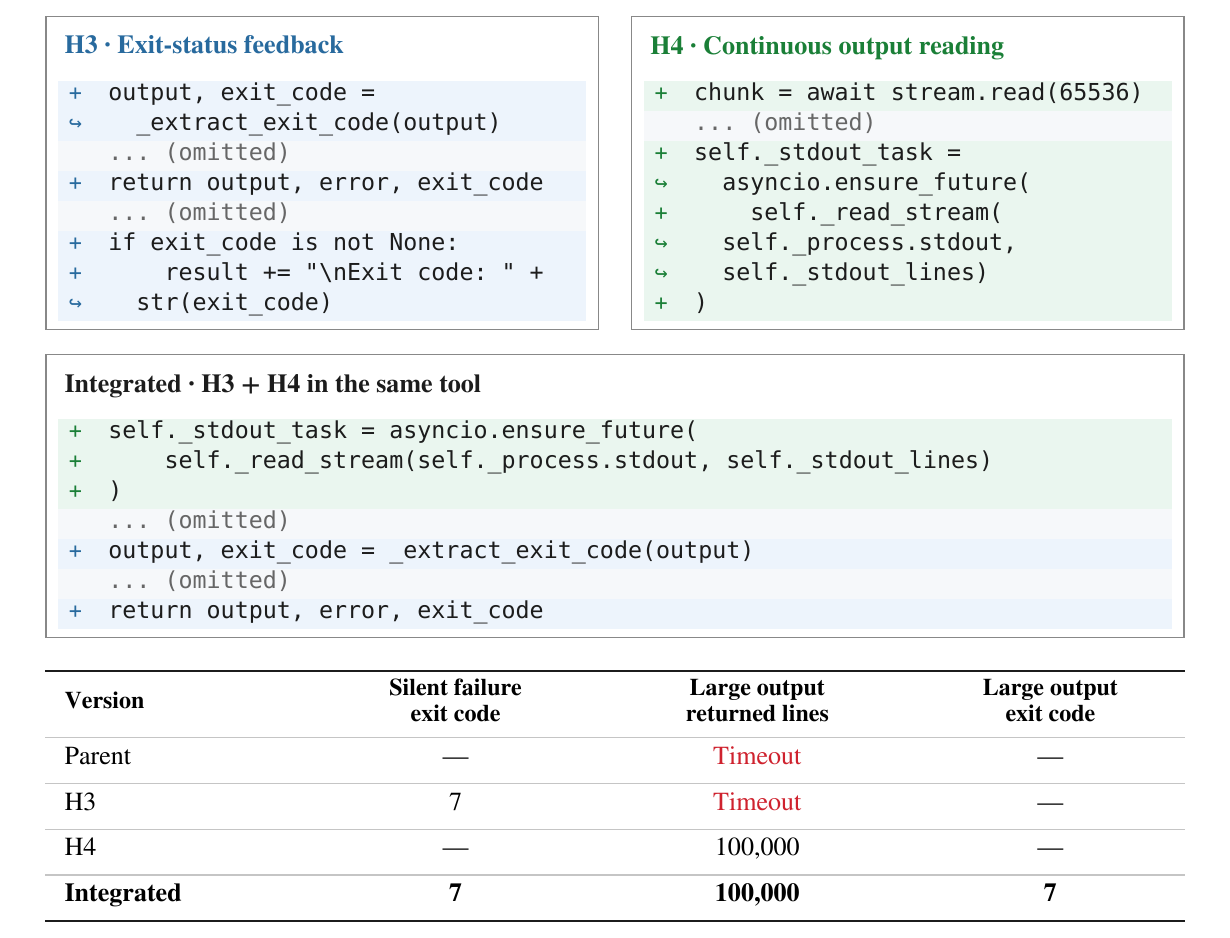}
    \caption{\textbf{Complementary code changes and their observed effects.}
    Matching blue and green highlights connect H3's exit-status feedback and
    H4's output reading to the integrated tool. The table compares all four versions
    on the same local probes; a dash denotes no returned exit status.}
    \label{fig:g5_combined_case}
\end{figure}

\paragraph{Repairing an exploratory implementation.}
Figure~\ref{fig:g1_combined_case} compares the G2-H6 manual tool-action parser
with its integrated counterpart. H6's brace matching skips only double-quoted
strings, so braces inside single-quoted commands can disrupt extraction.
Integration adds a helper that handles both quote styles. For the displayed
input containing \texttt{printf \{}, H6 extracts no valid action, whereas the
integrated parser recovers the command. The table shows the
resulting correction on three constructed edge cases while preserving the
two formats already handled by H6. These checks establish local parser behavior,
without attributing downstream score changes to this particular parsing path.

\begin{figure}[htbp]
    \centering
    \includegraphics[width=0.9\linewidth]{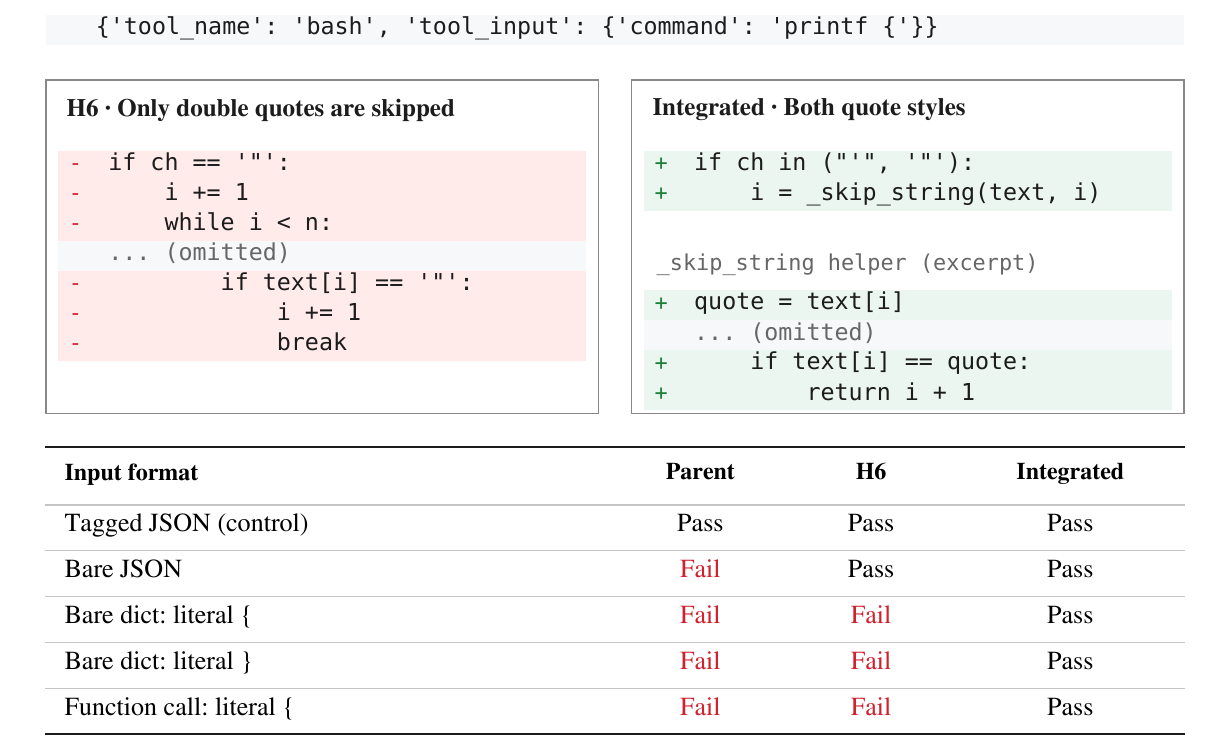}
    \caption{\textbf{An implementation refinement and five format checks.}
    Before/after excerpts reconstructed from archived H6 and integrated sources
    show the change in quote handling. The table compares the exact archived
    manual parsers on the same five inputs; parsed commands were not executed.}
    \label{fig:g1_combined_case}
\end{figure}

\paragraph{A review mechanism used during task solving.}
Figure~\ref{fig:nodebb_combined_case} connects the review code introduced by G9-H1
and retained during integration to an actual NodeBB trajectory. The added call
receives the exported patch and preceding solve history. During review, the agent
removes three Redis artifacts containing 15,710 added text lines while preserving
all three source-code diff blocks unchanged, reducing the patch from six files
to three. The review code is in \texttt{agents/universal\_coding\_agent.py}.
The final task is resolved; because
the pre-review patch has no separate official evaluation, the demonstrated
benefit is patch cleanup rather than a measured failure-to-success transition.

\begin{figure}[htbp]
    \centering
    \includegraphics[width=0.9\linewidth]{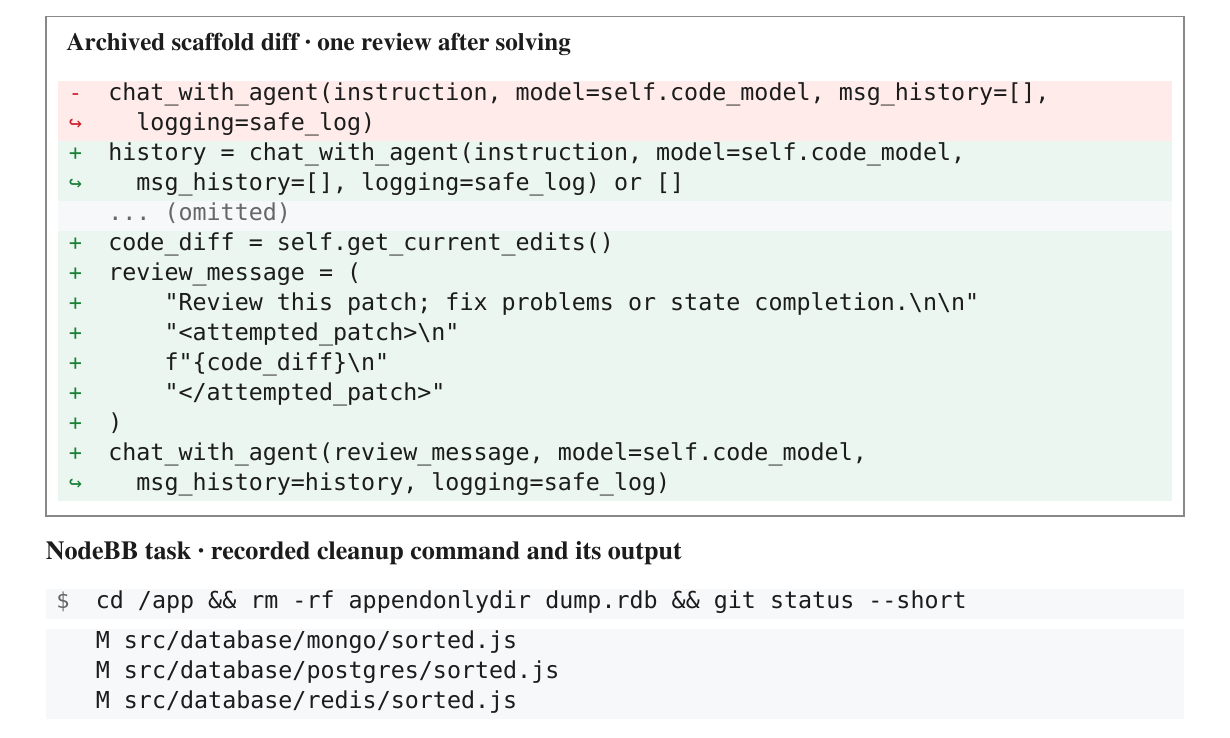}
    \caption{\textbf{From an evolved code change to observed task behavior.}
    The archived review patch and the recorded NodeBB cleanup command show how
    the added review removes non-source artifacts from the exported patch.}
    \label{fig:nodebb_combined_case}
\end{figure}

\paragraph{Reducing model-mediated test discovery.}
Figure~\ref{fig:qwen_g5_discovery} shows how G5 integration retains H7's deterministic test discovery while omitting H3's additional model-based planning stage. The archived H7 and integrated discovery methods are identical: they scan repository files and construct test commands, replacing the parent's \texttt{chat\_with\_agent} call. Local fixtures confirm file discovery and command construction for Python, JavaScript, Go, and Rust without model calls. This eliminates the model-mediated discovery stage; subsequent test execution remains agent-mediated.

\begin{figure}[htbp]
    \centering
    \includegraphics[width=0.9\linewidth]{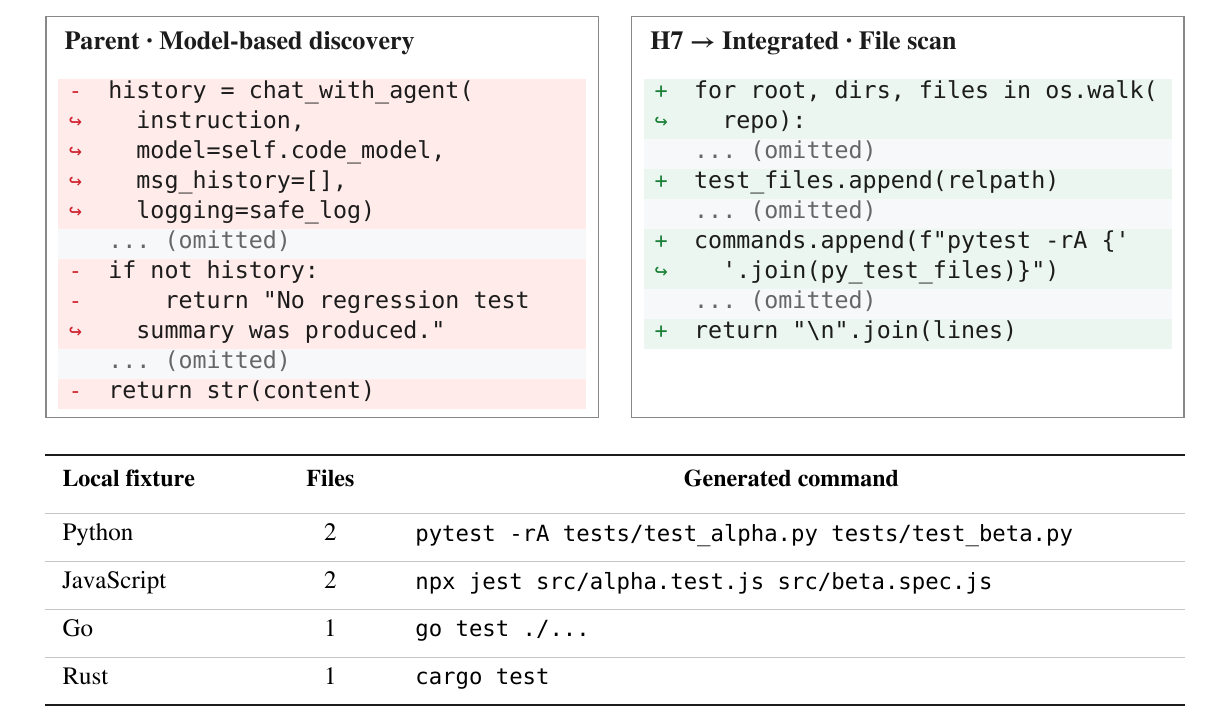}
    \caption{\textbf{Deterministic test discovery retained during Qwen integration.}
    The code change replaces model-mediated discovery with file scanning.
    H7 and the integrated version return the same outputs on four local fixtures;
    the displayed commands are generated, not executed.}
    \label{fig:qwen_g5_discovery}
\end{figure}
\clearpage

\paragraph{Composing and constraining edit recovery.}
Figure~\ref{fig:qwen_g8_editor} connects G8-H6's replacement fallback and G8-H4's immediate Python/JSON checks to their integrated implementation. Integration restricts H6's similarity-based matching to trailing-whitespace normalization. On the constructed inputs, H6 replaces \texttt{timeout\_ms} when asked to replace the absent \texttt{timeout\_s}; integration rejects this mismatch while retaining recovery from trailing-space differences. It also retains H4's checks on newly created Python and JSON files, which H6 does not check. The five local checks illustrate selective refinement of a child mechanism alongside composition. Verification reports errors after writing and does not roll back invalid edits.

\begin{figure}[htbp]
    \centering
    \includegraphics[width=0.9\linewidth]{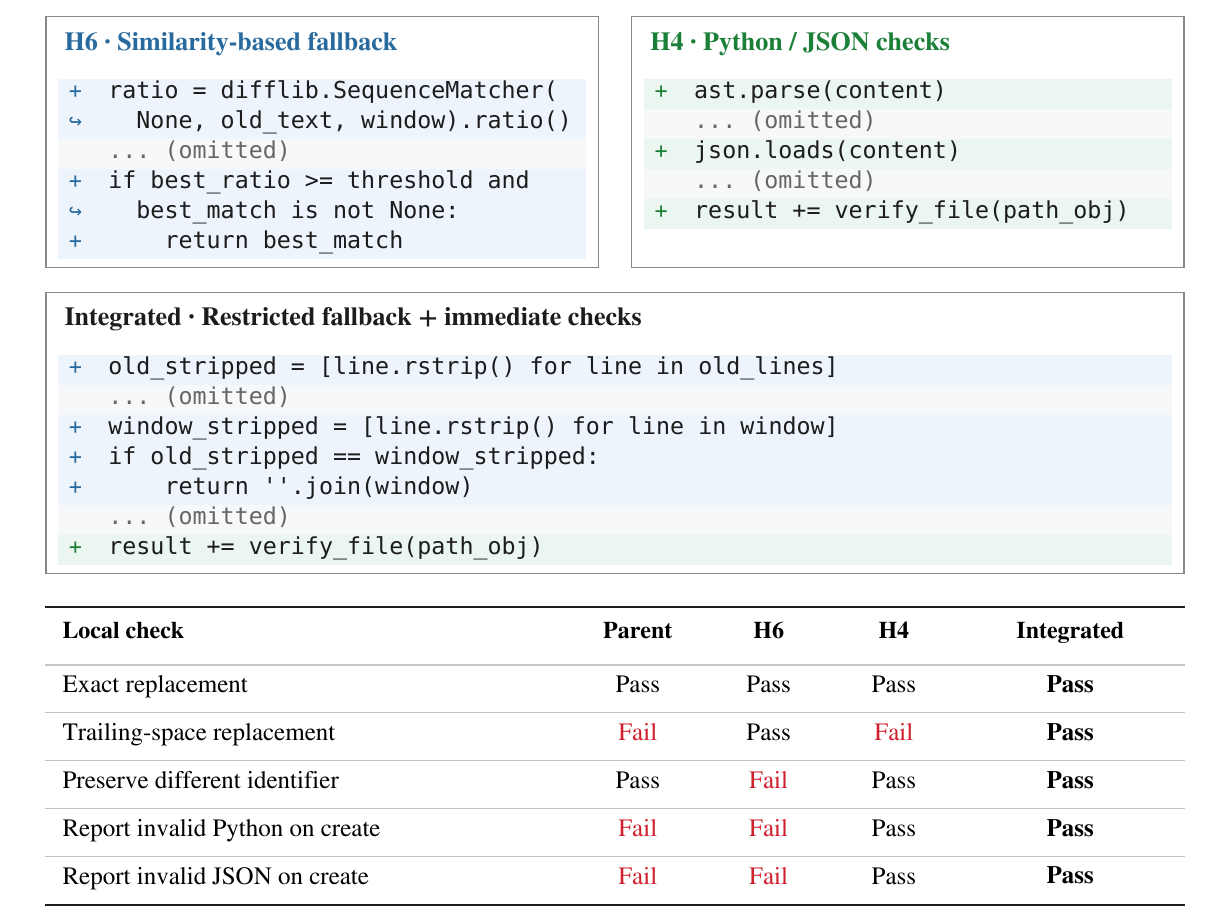}
    \caption{\textbf{Qwen integration combines edit recovery with immediate checks.}
    Blue traces H6's matching mechanism and its restricted integrated form;
    green traces H4's verification. Pass/Fail indicates the specified behavior
    on each constructed local input, rather than benchmark task success.}
    \label{fig:qwen_g8_editor}
\end{figure}
\clearpage

\paragraph{Using change-aware regression checks.}
Figure~\ref{fig:qwen_g9_task} traces G9-H1's unchanged test-selection code through a recorded Qutebrowser task involving dark-mode image classification. The mechanism identifies changed modules, locates tests that reference them, and falls back to all discovered tests when necessary. In this trajectory, it selects two of 176 test files and executes 141 passing test cases under a headless Qt configuration; the official evaluator also resolves the task. This shows that the evolved guidance is used during task solving without attributing final success or runtime reduction to this mechanism alone.

\begin{figure}[htbp]
    \centering
    \includegraphics[width=0.9\linewidth]{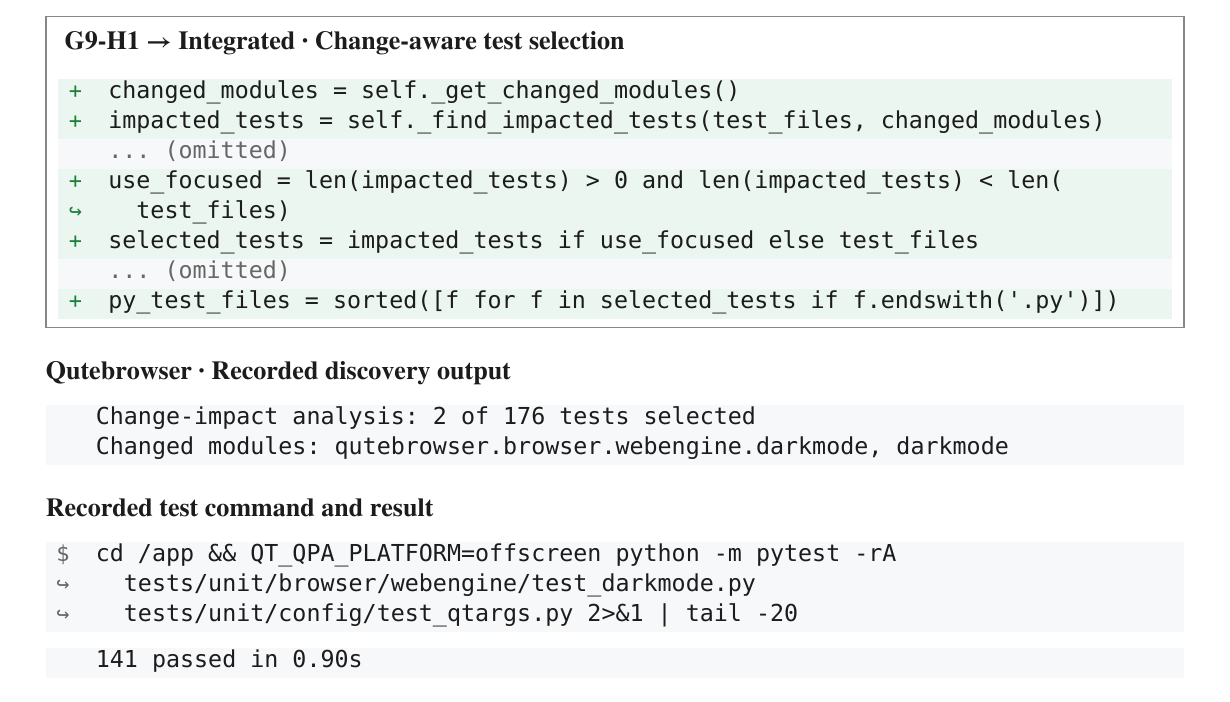}
    \caption{\textbf{From Qwen's evolved test selection to recorded task execution.}
    The retained code, discovery output, and executed command establish the
    connection.}
    \label{fig:qwen_g9_task}
\end{figure}

\newpage
\subsection{Initial Agent}
\label{app:initial_agent}

\subsubsection{Repository overview}

\begin{wrapfigure}{r}{0.3\textwidth}
    \centering
    \vspace{-18pt}
    \includegraphics[width=\linewidth]{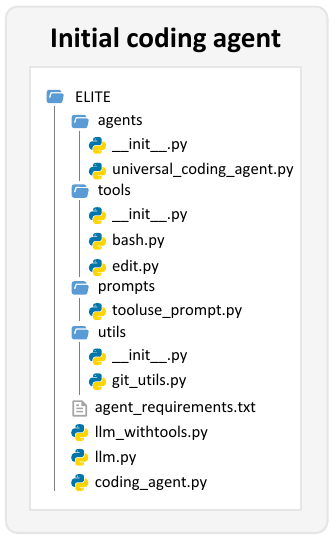}
    \caption{Initial coding agent repository.}
    \label{fig:initial_agent}
    \vspace{-38pt}
\end{wrapfigure}

The initial scaffold $\Sigma_0$ is an executable source-code
repository containing the agent's prompts, tools, utilities,
and control logic.
Given a task description and a working repository, the agent
uses an LLM to inspect the codebase, make changes, and perform
local checks.
It exposes two tools: \texttt{bash} executes shell commands,
and \texttt{editor} supports file inspection, creation, and
whole-file editing.
All methods start from the same scaffold for a given backbone,
and the same initial agent is evaluated across all three
benchmarks.
Figure~\ref{fig:initial_agent} shows the repository structure.

\subsubsection{Tools}
\label{app:initial_agent_tools}

The initial agent has two tools.
Together, they support repository inspection, code modification,
and local verification.

\paragraph{Bash.}
The \texttt{bash} tool accepts a shell command through its
\texttt{command} argument and executes it in the task environment.
The agent can use it to navigate and search the repository,
inspect dependencies and build configurations, run tests, and
review the resulting Git diff.
The implementation collects standard output and filtered standard
error and applies a 120-second command timeout.
Each invocation starts a new Bash session, so commands should
explicitly establish their working directory when necessary.
Changes to files remain available to subsequent tool calls.

\paragraph{Edit.}
The file-editing tool is implemented in \texttt{tools/edit.py}
and exposed to the model under the name \texttt{editor}.
It takes an absolute \texttt{path} and supports three commands.
The \texttt{view} command displays a file with line numbers or
lists a directory to a depth of two levels; file contents are
truncated after 10,000 characters before line numbers are added.
The \texttt{create} command writes the supplied
\texttt{file\_text} to a new file and rejects an existing path.
The \texttt{edit} command replaces the complete contents of an
existing file with \texttt{file\_text}.
This initial interface therefore performs file creation and
whole-file editing.

\subsubsection{Task Prompt}
\label{app:initial_agent_prompts}
The default execution path uses a brief system instruction to
establish a careful, generalist coding-agent role and encourage
tool use for inspection, editing, and testing.
Listing~\ref{lst:initial_task_prompt} presents the main task
template, shared across benchmarks, with the task description,
repository location, and optional context supplied at runtime.
Placeholder names are normalized for readability.

\par
\ifnum\value{WF@wrappedlines}>1\relax
    \vspace{\dimexpr\value{WF@wrappedlines}\baselineskip-\baselineskip\relax}
\fi
\WFclear

\begin{lstlisting}[
    style=eliteappendix,
    caption={Main task prompt of the initial coding agent.},
    label={lst:initial_task_prompt}
]
You are a generalist coding agent. Modify the repository at `{git_tempdir}` to solve the task below.

<task>
{problem_statement}
</task>

<repository_context>
{repository_context}
</repository_context>

<test_hint>
{test_description}
</test_hint>

Requirements:
1. Be dataset-agnostic: do not rely on benchmark names, task ids, hidden tests, or known leaderboard artifacts.
2. Be language-agnostic: inspect repository structure, dependency files, build files, and existing tests before deciding what to edit.
3. Prefer minimal, source-code-focused patches. Do not edit benchmark tests or create hard-coded task-id branches.
4. Use available tools to inspect files, run targeted tests, and edit code.
5. Before finishing, check `git diff` and ensure the patch is non-empty, relevant, and does not include accidental artifacts.
\end{lstlisting}

\subsection{Prompts Used by ELITE}
\label{app:elite_prompts}

This section presents the templates for hypothesis generation,
exploratory self-modification, and integration.
Braced placeholders denote inputs supplied at runtime.
Hypothesis generation uses separate system and user messages;
exploration and integration supply self-modification tasks to
the coding agent described in Appendix~\ref{app:initial_agent}.

\subsubsection{Improvement Hypothesis Generation}
\label{app:hypothesis_generation_prompts}

\paragraph{System prompt.}
Listing~\ref{lst:hypothesis_generation_system} specifies the
objectives and constraints for jointly generating focused,
distinct, and complementary improvement hypotheses grounded
in the current implementation.

\begin{lstlisting}[
    style=eliteappendix,
    caption={System prompt for improvement hypothesis generation.},
    label={lst:hypothesis_generation_system}
]
You design one generation of focused exploratory mutations for a self-evolving coding agent.

Given the exact current runtime source and a compact evolution report, generate the requested number of hypotheses jointly. The set should contain the most promising, distinct, and complementary behavioral improvements visible in the current implementation.

Each hypothesis must:
- address one concrete behavioral limitation;
- propose one focused mechanism that can be implemented independently;
- be complete enough to produce a meaningful child, but not bundle unrelated features;
- modify only the runtime components necessary for that mechanism;
- remain grounded in the supplied source rather than a fixed category list.

Across the set, avoid near-duplicates and unnecessary concentration in one subsystem. Do not treat tools, model wrappers, prompts, memory, planning, or new modules as inherently higher-impact than existing mechanisms. Do not generate extra candidates for a later selection stage.

Do not use benchmark IDs, hidden tests, reference solutions, leaderboard artifacts, or benchmark-specific parsers. Return one JSON object only.
\end{lstlisting}

\paragraph{User prompt.}
Listing~\ref{lst:hypothesis_generation_user} supplies the
current runtime source, a compact evolution report, and the
required JSON schema.
Each hypothesis describes a capability objective, a limitation,
a proposed mechanism, acceptance conditions, and its
distinction from the other hypotheses.
Structurally invalid responses trigger a request for corrected
JSON.

\begin{lstlisting}[
    style=eliteappendix,
    caption={User prompt for improvement hypothesis generation.},
    label={lst:hypothesis_generation_user}
]
## Compact evolution report
{evolution_report}

## Exact current runtime source
{parent_source}

Generate exactly {hypothesis_count} focused, distinct, and complementary hypotheses. Return this schema:

{
  "hypotheses": [
    {
      "hypothesis_id": "H1",
      "capability": "Behavioral capability to improve",
      "limitation": "Concrete limitation visible in the current source",
      "mechanism": "One focused implementation mechanism",
      "acceptance": "Observable conditions showing the mechanism is complete",
      "distinctive_basis": "Why this hypothesis is meaningfully different from the others"
    }
  ]
}

Requirements:
- Include exactly {hypothesis_count} hypotheses, identified H1 through H{hypothesis_count}.
- Choose the final set directly; there is no later curation stage.
- Each hypothesis must represent one coherent intervention, not a broad redesign or a tiny cosmetic edit.
- Make the hypotheses complementary enough that a later synthesis step can combine compatible successes.
- Keep every field concise.
\end{lstlisting}

\subsubsection{Exploratory Self-Modification}
\label{app:exploratory_modification_prompt}

\paragraph{Exploration task prompt.}
Listing~\ref{lst:exploratory_modification} converts each
hypothesis into an executable self-modification task.
The capability objective, limitation, mechanism, and acceptance
conditions are inserted into the corresponding placeholders.
Each exploratory child starts from the same parent and
implements one focused intervention.

\begin{lstlisting}[
    style=eliteappendix,
    caption={Task prompt for implementing an exploratory self-modification.},
    label={lst:exploratory_modification}
]
# Focused Self-Modification Task

## Capability Objective
{capability}

## Current Limitation
{limitation}

## Mechanism to Implement
{mechanism}

## Acceptance Conditions
{acceptance}

## Scope
Implement exactly one focused behavioral intervention. Touch the smallest coherent set of runtime components needed to make the mechanism complete and usable on the default solve path. Reuse existing interfaces where possible. Do not add adjacent features or broadly redesign the agent.
Do not modify tests, benchmark/evaluation infrastructure, proxy/evolution logic, run artifacts, or encode benchmark-specific information.
\end{lstlisting}

\subsubsection{Integration}
\label{app:integration_prompt}

\paragraph{Integration task prompt.}
Listing~\ref{lst:integration_task} guides the construction of
one integrated descendant from the unmodified common parent.
The input manifest provides candidate descriptions,
deterministic check results, and references to complete patches.
The integrator considers all candidates at the manifest level,
initially shortlists one to three compatible children, inspects
their complete patches, and reimplements selected mechanisms
as a coherent descendant.
Additional patches are inspected only when needed to supply a
missing compatible mechanism.
Downstream task outcomes are not provided to the integrator.

\begin{lstlisting}[
    style=eliteappendix,
    caption={Task prompt for integrating exploratory modifications.},
    label={lst:integration_task}
]
# Integrated Self-Modification Task

This is the synthesis step of the current evolution generation.

The repository currently contains the unmodified parent agent. Before changing code, read:

    {manifest_path}

The manifest contains compact evidence for all {child_count} exploratory children and exact bounded patch paths.

Use this sequence:
1. Consider every child from the manifest-level evidence.
2. Ignore empty children and children that failed deterministic checks.
3. Shortlist one to three compatible children with the strongest behavioral mechanisms.
4. For each shortlisted child, read every file in its `patch_chunks` list, in order. This reconstructs its complete patch without a large clipped tool response.
5. Begin implementation after the shortlist has been fully inspected. Read another child's full patch only when its manifest evidence identifies a missing, compatible mechanism.

Create one coherent integrated child that captures the strongest compatible capability improvements discovered across the exploratory children.

Requirements:
- Consider every exploratory child at manifest level, but do not mechanically merge every patch.
- Keep useful mechanisms; reject redundant, conflicting, incomplete, speculative, or disconnected parts.
- Reimplement selected ideas as one consistent design starting from the current parent repository.
- Preserve a clear default solve path. Reuse existing mechanisms when they already provide the required behavior; any new component or control-flow change must be necessary, reachable, and fully integrated.
- Prefer a small number of mutually reinforcing capabilities over an incoherent bundle. Do not favor infrastructure changes merely because several children modify the same central subsystem.
- Keep the default solve path bounded: do not stack multiple full-agent passes or open-ended repair loops. Added stages must replace or simplify existing work rather than multiply end-to-end model calls.
- Do not spend the tool budget exhaustively reading low-value or invalid patches; move to implementation once the shortlist is understood.
- Before finishing, inspect `git diff` and ensure the resulting patch is non-empty and internally consistent.
- Patch size and file count are neither rewards nor penalties.
- Do not modify tests, benchmark/evaluation infrastructure, proxy/evolution logic, run artifacts, or encode benchmark-specific information.

The result of this task is the single integrated child proposed for this generation.
\end{lstlisting}

\subsection{Benchmarks and Task Subsets}
\label{app:benchmark_details}

We evaluate the agents on fixed subsets of SWE-bench Pro, SWE-bench Verified, and Aider Polyglot, containing 60, 50, and 50 tasks, respectively. SWE-bench Pro is used to provide evolutionary feedback for the baselines, whereas SWE-bench Verified and Polyglot are used only for post-evolution transfer evaluation and never provide feedback during ELITE evolution. The benchmark referred to as Verified in the main results is specifically SWE-bench Verified. The task identifiers are listed in Tables~\ref{tab:pro_task_ids}--\ref{tab:polyglot_task_ids}.

\paragraph{Task execution.}
Each task is run in a separate container with a fresh agent
process and the benchmark-specified repository state.
An agent run may include multiple model responses, tool calls,
and local tests.
The resulting code changes are assessed by the corresponding
benchmark evaluator.

\paragraph{Evolution feedback.}
DGM and HGM use SWE-bench Pro task outcomes during evolution.
ELITE does not use downstream task outcomes from evolved
candidates to decide integration or the next-agent update.
Running available tests while solving an individual task is
distinct from returning benchmark evaluation results to the
outer self-improvement procedure.

\subsubsection{SWE-bench Pro}
\label{app:swebench_pro_details}

SWE-bench Pro consists of repository-level software engineering tasks that require modifying an existing codebase.\footnote{
    \url{https://huggingface.co/datasets/ScaleAI/SWE-bench_Pro}
}
The agent receives the problem statement and, when available, additional requirements, interface specifications, and test hints. Its generated patch is assessed using the SWE-bench Pro evaluation harness. We use a subset of 60 tasks spanning 11 repositories for the
main evolution comparison. Table~\ref{tab:pro_task_ids} lists these tasks by repository. Each complete identifier is obtained by concatenating the group prefix, the hash component, and the suffix.

\begin{table}[p]
\centering
\caption{
The 60 SWE-bench Pro task IDs, grouped by repository.
Each ID is the concatenation of the group prefix, the hash
component, and the suffix.
An em dash denotes an empty suffix.
}
\label{tab:pro_task_ids}
\scriptsize
\setlength{\tabcolsep}{7pt}
\renewcommand{\arraystretch}{0.92}
\begin{tabularx}{\linewidth}{
@{}
>{\raggedright\arraybackslash}X
>{\raggedright\arraybackslash}X
@{}
}
\toprule
Hash component & Suffix \\
\midrule

\addlinespace[1.5pt]
\multicolumn{2}{@{}l@{}}{
\texttt{instance\_ansible\_\_ansible-}\quad (3 tasks)} \\
\texttt{de5858f48dc9e1ce9117034e0d7e76806f420ca8} & \texttt{-v1055803c3a812189a1133297f7f5468579283f86} \\
\texttt{0fd88717c953b92ed8a50495d55e630eb5d59166} & \texttt{-vba6da65a0f3baefda7a058ebbd0a8dcafb8512f5} \\
\texttt{cd473dfb2fdbc97acf3293c134b21cbbcfa89ec3} & \texttt{-vba6da65a0f3baefda7a058ebbd0a8dcafb8512f5} \\

\addlinespace[1.5pt]
\multicolumn{2}{@{}l@{}}{
\texttt{instance\_element-hq\_\_element-web-}\quad (5 tasks)} \\
\texttt{5e8488c2838ff4268f39db4a8cca7d74eecf5a7e} & \texttt{-vnan} \\
\texttt{66d0b318bc6fee0d17b54c1781d6ab5d5d323135} & \texttt{-vnan} \\
\texttt{18c03daa865d3c5b10e52b669cd50be34c67b2e5} & \texttt{-vnan} \\
\texttt{e15ef9f3de36df7f318c083e485f44e1de8aad17} & \textemdash \\
\texttt{ca58617cee8aa91c93553449bfdf9b3465a5119b} & \texttt{-vnan} \\

\addlinespace[1.5pt]
\multicolumn{2}{@{}l@{}}{
\texttt{instance\_flipt-io\_\_flipt-}\quad (7 tasks)} \\
\texttt{abaa5953795afb9c621605bb18cb32ac48b4508c} & \textemdash \\
\texttt{3d5a345f94c2adc8a0eaa102c189c08ad4c0f8e8} & \textemdash \\
\texttt{c6a7b1fd933e763b1675281b30077e161fa115a1} & \textemdash \\
\texttt{2ca5dfb3513e4e786d2b037075617cccc286d5c3} & \textemdash \\
\texttt{507170da0f7f4da330f6732bffdf11c4df7fc192} & \textemdash \\
\texttt{292fdaca9be39e6a921aaa8874c011d0fdd3e874} & \textemdash \\
\texttt{0b119520afca1cf25c470ff4288c464d4510b944} & \textemdash \\

\addlinespace[1.5pt]
\multicolumn{2}{@{}l@{}}{
\texttt{instance\_future-architect\_\_vuls-}\quad (2 tasks)} \\
\texttt{f0b3a8b1db98eb1bd32685f1c36c41a99c3452ed} & \textemdash \\
\texttt{e52fa8d6ed1d23e36f2a86e5d3efe9aa057a1b0d} & \textemdash \\

\addlinespace[1.5pt]
\multicolumn{2}{@{}l@{}}{
\texttt{instance\_gravitational\_\_teleport-}\quad (3 tasks)} \\
\texttt{37c3724d0d6637e959e39408ee351565d73afe71} & \texttt{-vee9b09fb20c43af7e520f57e9239bbcf46b7113d} \\
\texttt{65438e6e44b6ce51458d09b7bb028a2797cfb0ea} & \texttt{-vce94f93ad1030e3136852817f2423c1b3ac37bc4} \\
\texttt{0ac7334939981cf85b9591ac295c3816954e287e} & \textemdash \\

\addlinespace[1.5pt]
\multicolumn{2}{@{}l@{}}{
\texttt{instance\_internetarchive\_\_openlibrary-}\quad (3 tasks)} \\
\texttt{8a5a63af6e0be406aa6c8c9b6d5f28b2f1b6af5a} & \texttt{-v0f5aece3601a5b4419f7ccec1dbda2071be28ee4} \\
\texttt{3f7db6bbbcc7c418b3db72d157c6aed1d45b2ccf} & \texttt{-v430f20c722405e462d9ef44dee7d34c41e76fe7a} \\
\texttt{798a582540019363d14b2090755cc7b89a350788} & \texttt{-v430f20c722405e462d9ef44dee7d34c41e76fe7a} \\

\addlinespace[1.5pt]
\multicolumn{2}{@{}l@{}}{
\texttt{instance\_navidrome\_\_navidrome-}\quad (3 tasks)} \\
\texttt{874b17b8f614056df0ef021b5d4f977341084185} & \textemdash \\
\texttt{669c8f4c49a7ef51ac9a53c725097943f67219eb} & \textemdash \\
\texttt{89b12b34bea5687c70e4de2109fd1e7330bb2ba2} & \textemdash \\

\addlinespace[1.5pt]
\multicolumn{2}{@{}l@{}}{
\texttt{instance\_NodeBB\_\_NodeBB-}\quad (5 tasks)} \\
\texttt{0f788b8eaa4bba3c142d171fd941d015c53b65fc} & \texttt{-v0ec6d6c2baf3cb4797482ce4829bc25cd5716649} \\
\texttt{b1f9ad5534bb3a44dab5364f659876a4b7fe34c1} & \texttt{-vnan} \\
\texttt{cfc237c2b79d8c731bbfc6cadf977ed530bfd57a} & \texttt{-v0495b863a912fbff5749c67e860612b91825407c} \\
\texttt{087e6020e490b4a1759f38c1ad03869511928263} & \texttt{-vf2cf3cbd463b7ad942381f1c6d077626485a1e9e} \\
\texttt{da0211b1a001d45d73b4c84c6417a4f1b0312575} & \texttt{-vf2cf3cbd463b7ad942381f1c6d077626485a1e9e} \\

\addlinespace[1.5pt]
\multicolumn{2}{@{}l@{}}{
\texttt{instance\_protonmail\_\_webclients-}\quad (5 tasks)} \\
\texttt{b9387af4cdf79c2cb2a221dea33d665ef789512e} & \textemdash \\
\texttt{6f8916fbadf1d1f4a26640f53b5cf7f55e8bedb7} & \textemdash \\
\texttt{b387b24147e4b5ec3b482b8719ea72bee001462a} & \textemdash \\
\texttt{f080ffc38e2ad7bddf2e93e5193e82c20c7a11e7} & \textemdash \\
\texttt{cb8cc309c6968b0a2a5fe4288d0ae0a969ff31e1} & \textemdash \\

\addlinespace[1.5pt]
\multicolumn{2}{@{}l@{}}{
\texttt{instance\_qutebrowser\_\_qutebrowser-}\quad (9 tasks)} \\
\texttt{de4a1c1a2839b5b49c3d4ce21d39de48d24e2091} & \texttt{-v2ef375ac784985212b1805e1d0431dc8f1b3c171} \\
\texttt{8cd06741bb56cdca49f5cdc0542da97681154315} & \texttt{-v5149fcda2a9a6fe1d35dfed1bade1444a11ef271} \\
\texttt{c09e1439f145c66ee3af574386e277dd2388d094} & \texttt{-v2ef375ac784985212b1805e1d0431dc8f1b3c171} \\
\texttt{fcfa069a06ade76d91bac38127f3235c13d78eb1} & \texttt{-v5fc38aaf22415ab0b70567368332beee7955b367} \\
\texttt{9ed748effa8f3bcd804612d9291da017b514e12f} & \texttt{-v363c8a7e5ccdf6968fc7ab84a2053ac78036691d} \\
\texttt{70248f256f93ed9b1984494d0a1a919ddd774892} & \texttt{-v2ef375ac784985212b1805e1d0431dc8f1b3c171} \\
\texttt{ec2dcfce9eee9f808efc17a1b99e227fc4421dea} & \texttt{-v5149fcda2a9a6fe1d35dfed1bade1444a11ef271} \\
\texttt{cc360cd4a34a126274c7b51f3b63afbaf3e05a02} & \texttt{-v5fc38aaf22415ab0b70567368332beee7955b367} \\
\texttt{fd6790fe8c02b144ab2464f1fc8ab3d02ce3c476} & \texttt{-v2ef375ac784985212b1805e1d0431dc8f1b3c171} \\

\addlinespace[1.5pt]
\multicolumn{2}{@{}l@{}}{
\texttt{instance\_tutao\_\_tutanota-}\quad (15 tasks)} \\
\texttt{befce4b146002b9abc86aa95f4d57581771815ce} & \texttt{-vee878bb72091875e912c52fc32bc60ec3760227b} \\
\texttt{40e94dee2bcec2b63f362da283123e9df1874cc1} & \texttt{-vc4e41fd0029957297843cb9dec4a25c7c756f029} \\
\texttt{219bc8f05d7b980e038bc1524cb021bf56397a1b} & \texttt{-vee878bb72091875e912c52fc32bc60ec3760227b} \\
\texttt{fe240cbf7f0fdd6744ef7bef8cb61676bcdbb621} & \texttt{-vc4e41fd0029957297843cb9dec4a25c7c756f029} \\
\texttt{8513a9e8114a8b42e64f4348335e0f23efa054c4} & \texttt{-vee878bb72091875e912c52fc32bc60ec3760227b} \\
\texttt{db90ac26ab78addf72a8efaff3c7acc0fbd6d000} & \texttt{-vbc0d9ba8f0071fbe982809910959a6ff8884dbbf} \\
\texttt{b4934a0f3c34d9d7649e944b183137e8fad3e859} & \texttt{-vbc0d9ba8f0071fbe982809910959a6ff8884dbbf} \\
\texttt{d1aa0ecec288bfc800cfb9133b087c4f81ad8b38} & \texttt{-vbc0d9ba8f0071fbe982809910959a6ff8884dbbf} \\
\texttt{de49d486feef842101506adf040a0f00ded59519} & \texttt{-v10a26bfb45a064b93f4fc044a0254925037b88f1} \\
\texttt{1e516e989b3c0221f4af6b297d9c0e4c43e4adc3} & \texttt{-vbc0d9ba8f0071fbe982809910959a6ff8884dbbf} \\
\texttt{1ff82aa365763cee2d609c9d19360ad87fdf2ec7} & \texttt{-vc4e41fd0029957297843cb9dec4a25c7c756f029} \\
\texttt{09c2776c0fce3db5c6e18da92b5a45dce9f013aa} & \texttt{-vbc0d9ba8f0071fbe982809910959a6ff8884dbbf} \\
\texttt{51818218c6ae33de00cbea3a4d30daac8c34142e} & \texttt{-vc4e41fd0029957297843cb9dec4a25c7c756f029} \\
\texttt{f373ac3808deefce8183dad8d16729839cc330c1} & \texttt{-v2939aa9f4356f0dc9f523ee5ce19d09e08ab979b} \\
\texttt{fbdb72a2bd39b05131ff905780d9d4a2a074de26} & \texttt{-vbc0d9ba8f0071fbe982809910959a6ff8884dbbf} \\

\bottomrule
\end{tabularx}
\end{table}

\subsubsection{SWE-bench Verified}
\label{app:swebench_verified_details}

SWE-bench Verified is a human-validated subset of SWE-bench containing 500 repository-level tasks derived from real GitHub issues.\footnote{
    \url{https://openai.com/index/introducing-swe-bench-verified/}
}
The agent receives an issue description and modifies the corresponding repository. The evaluator checks whether the issue is resolved and whether the required existing tests continue to pass. We follow DGM \cite{zhang2026darwin} and use the same 50-task subset, comprising 25 Django tasks and 25 Sphinx tasks. Table~\ref{tab:verified_task_ids} lists the task numbers
under their respective repository prefixes.

\begin{table}[htbp]
\centering
\caption{
The 50 SWE-bench Verified task IDs.
Each ID consists of the prefix in the first column followed
by one number from the second column.
}
\label{tab:verified_task_ids}
\small
\setlength{\tabcolsep}{4pt}
\renewcommand{\arraystretch}{1.1}
\begin{tabularx}{\linewidth}{
@{}
>{\raggedright\arraybackslash}p{0.25\linewidth}
>{\raggedright\arraybackslash}X
@{}
}
\toprule
ID prefix & Task numbers \\
\midrule

\texttt{django\_\_django-} &
\mbox{\texttt{9296}},
\mbox{\texttt{11790}},
\mbox{\texttt{11815}},
\mbox{\texttt{11848}},
\mbox{\texttt{11880}},
\mbox{\texttt{11885}},
\mbox{\texttt{11951}},
\mbox{\texttt{11964}},
\mbox{\texttt{11999}},
\mbox{\texttt{12039}},
\mbox{\texttt{12050}},
\mbox{\texttt{12143}},
\mbox{\texttt{12155}},
\mbox{\texttt{12193}},
\mbox{\texttt{12209}},
\mbox{\texttt{12262}},
\mbox{\texttt{12273}},
\mbox{\texttt{12276}},
\mbox{\texttt{12304}},
\mbox{\texttt{12308}},
\mbox{\texttt{12325}},
\mbox{\texttt{12406}},
\mbox{\texttt{12708}},
\mbox{\texttt{12713}},
\mbox{\texttt{12774}} \\

\texttt{sphinx-doc\_\_sphinx-} &
\mbox{\texttt{7454}},
\mbox{\texttt{7590}},
\mbox{\texttt{7748}},
\mbox{\texttt{7757}},
\mbox{\texttt{7985}},
\mbox{\texttt{8035}},
\mbox{\texttt{8056}},
\mbox{\texttt{8265}},
\mbox{\texttt{8269}},
\mbox{\texttt{8475}},
\mbox{\texttt{8548}},
\mbox{\texttt{8551}},
\mbox{\texttt{8638}},
\mbox{\texttt{8721}},
\mbox{\texttt{9229}},
\mbox{\texttt{9230}},
\mbox{\texttt{9281}},
\mbox{\texttt{9320}},
\mbox{\texttt{9367}},
\mbox{\texttt{9461}},
\mbox{\texttt{9698}},
\mbox{\texttt{10449}},
\mbox{\texttt{10466}},
\mbox{\texttt{10673}},
\mbox{\texttt{11510}} \\

\bottomrule
\end{tabularx}
\end{table}

\subsubsection{Polyglot}
\label{app:polyglot_details}

The Aider Polyglot benchmark contains 225 programming
exercises drawn from Exercism and covers C++, Go, Java,
JavaScript, Python, and Rust.\footnote{
    \url{https://aider.chat/2024/12/21/polyglot.html}
}
The agent edits the provided solution files to implement the requested behavior. We follow DGM \cite{zhang2026darwin} and use the same 50-task subset for transfer evaluation, including 4 C++ tasks, 8 Go tasks, 11 Java tasks,
8 JavaScript tasks, 10 Python tasks, and 9 Rust tasks. Table~\ref{tab:polyglot_task_ids} lists the exercises under their language prefixes. Each task is attempted in one agent run, which may include tool use and local testing. After the run, the harness retains the modified solution files, restores the benchmark test revision, and executes the language-specific evaluation command. A task is marked as resolved when this command exits successfully. Final evaluation feedback is not followed by an additional agent repair attempt.

\begin{table}[htbp]
\centering
\caption{
The 50 Polyglot task IDs, grouped by language.
Each ID consists of the prefix in the first column followed
by one exercise name from the second column.
}
\label{tab:polyglot_task_ids}
\small
\setlength{\tabcolsep}{4pt}
\renewcommand{\arraystretch}{1.1}
\begin{tabularx}{\linewidth}{
@{}
>{\raggedright\arraybackslash}p{0.18\linewidth}
>{\raggedright\arraybackslash}X
@{}
}
\toprule
ID prefix & Exercise names \\
\midrule

\texttt{cpp\_\_} &
\mbox{\texttt{bank-account}},
\mbox{\texttt{diamond}},
\mbox{\texttt{gigasecond}},
\mbox{\texttt{queen-attack}} \\

\texttt{go\_\_} &
\mbox{\texttt{connect}},
\mbox{\texttt{counter}},
\mbox{\texttt{crypto-square}},
\mbox{\texttt{error-handling}},
\mbox{\texttt{food-chain}},
\mbox{\texttt{markdown}},
\mbox{\texttt{transpose}},
\mbox{\texttt{word-search}} \\

\texttt{java\_\_} &
\mbox{\texttt{bowling}},
\mbox{\texttt{custom-set}},
\mbox{\texttt{forth}},
\mbox{\texttt{go-counting}},
\mbox{\texttt{ledger}},
\mbox{\texttt{mazy-mice}},
\mbox{\texttt{react}},
\mbox{\texttt{rest-api}},
\mbox{\texttt{satellite}},
\mbox{\texttt{wordy}},
\mbox{\texttt{zipper}} \\

\texttt{javascript\_\_} &
\mbox{\texttt{forth}},
\mbox{\texttt{grade-school}},
\mbox{\texttt{ocr-numbers}},
\mbox{\texttt{queen-attack}},
\mbox{\texttt{say}},
\mbox{\texttt{simple-linked-list}},
\mbox{\texttt{triangle}},
\mbox{\texttt{wordy}} \\

\texttt{python\_\_} &
\mbox{\texttt{dominoes}},
\mbox{\texttt{dot-dsl}},
\mbox{\texttt{food-chain}},
\mbox{\texttt{grade-school}},
\mbox{\texttt{poker}},
\mbox{\texttt{scale-generator}},
\mbox{\texttt{sgf-parsing}},
\mbox{\texttt{tree-building}},
\mbox{\texttt{wordy}},
\mbox{\texttt{zipper}} \\

\texttt{rust\_\_} &
\mbox{\texttt{accumulate}},
\mbox{\texttt{doubly-linked-list}},
\mbox{\texttt{gigasecond}},
\mbox{\texttt{macros}},
\mbox{\texttt{pig-latin}},
\mbox{\texttt{poker}},
\mbox{\texttt{react}},
\mbox{\texttt{say}},
\mbox{\texttt{wordy}} \\

\bottomrule
\end{tabularx}
\end{table}

\subsection{Experimental Settings}
\label{app:experimental_settings}

\subsubsection{Evolution Configuration}

\paragraph{Shared model settings.}
We use DeepSeek-V4-Flash and Qwen-3.7-Plus as LLM  backbones. All methods share the same initial scaffold for a given backbone. The maximum output budget per model response is 16,384 tokens (\texttt{MAX\_OUTPUT\_TOKENS=16384}); hypothesis generation uses the same budget (\texttt{PORTFOLIO\_MAX\_TOKENS=16384}). Thinking is enabled through \texttt{DEEPSEEK\_THINKING=enabled} for DeepSeek and \texttt{QWEN\_ENABLE\_THINKING=1} for Qwen.

\paragraph{Controlled execution environment.}
Agent execution and deterministic validity checks run inside isolated \texttt{Docker} sandbox containers built from the experiment runtime image. Each exploratory or integrated child is executed in a separate containerized working copy with the timeout shown in Table~\ref{tab:evolution_hyperparameters}; containers are removed after the attempt; the host process handles orchestration and records status and artifacts. This containment isolates child mutations and tool execution from the host environment.

\paragraph{Evolution settings.}
Table~\ref{tab:evolution_hyperparameters} summarizes the main search parameters. Parallel workers denotes the configured concurrency limit. ELITE generates $K=7$ exploratory children per valid hypothesis portfolio for both backbones. It uses eight workers for both backbones; the maximum exploration concurrency is therefore $\min(8,7)=7$. Integration follows exploration. HGM uses the same search configuration for both backbones, with 20 parallel workers shared by expansion and evaluation.

\begin{table}[htbp]
\centering
\small
\setlength{\tabcolsep}{5pt}
\renewcommand{\arraystretch}{1.06}
\caption{\textbf{Main evolution parameters.}}
\label{tab:evolution_hyperparameters}
\begin{tabular*}{\linewidth}{@{\extracolsep{\fill}}lll@{}}
\toprule
\textbf{Method} & \textbf{Parameter} & \textbf{Value} \\
\midrule
\method & Max generations, $T$ & 10 \\
 & Exploration width, $K$ & 7 \\
 & Parallel workers & 8 \\
 & Exploration timeout & 1,800 s \\
 & Integration timeout & 1,800 s \\
\midrule
DGM & Candidates per generation & 2 \\
 & Parallel workers & 1 \\
 & Evaluation parallel workers & 8 \\
 & Self-modification timeout & 1,800 s \\
\midrule
HGM & Task-evaluation budget & 800 \\
 & Parallel workers & 20 \\
 & Initial expansions & 5 \\
 & $\alpha$, $\beta$ & 0.6, 1.0 \\
 & Cool-down & Enabled \\
 & Final percentile & 1.0 \\
 & Self-modification timeout & 3,600 s \\
 & Evaluation timeout & 3,600 s \\
\bottomrule
\end{tabular*}
\end{table}

\paragraph{Search budgets.}
Following HGM's evaluation-budget convention~\citep{wang2026huxley}, HGM is configured for 800 task evaluations, and DGM is compared over a search prefix with approximately the same evaluation count. ELITE instead uses a fixed ten-generation horizon, since it performs no online benchmark evaluation of new descendants. These are method-specific search budgets, not matched compute budgets.

\paragraph{Runtime comparison.}
All methods run on the same server using the execution settings in Table~\ref{tab:evolution_hyperparameters}. Search concurrency and task-evaluation concurrency operate at different levels. Under the settings in Table~\ref{tab:evolution_hyperparameters}, DGM uses one self-modification worker, but evaluates each candidate's benchmark tasks with up to eight parallel workers. HGM follows the official 20-worker configuration, sharing its pool between expansion and evaluation. \method{} uses an eight-worker pool for $K=7$ same-parent exploratory children, followed by integration; its exploration concurrency is therefore at most seven, even if the pool limit is raised to 20. Increasing candidate counts to occupy additional workers would change search breadth, while identical worker limits alone would not ensure equal active concurrency or resource use. We therefore report practical turnaround for the evaluated configurations and stopping budgets, including search-time task evaluation where applicable. The time ratios remain configuration-dependent; task-evaluation counts and billed API expenditure provide complementary measures of evolution cost.

\paragraph{Repetitions and evaluation budget.}
We report three independent evolution runs for \method{} with each backbone and for DGM and HGM with DeepSeek; the Qwen baselines use one run each. Repeating an evaluation-driven baseline requires rerunning the entire search, including approximately 800 task evaluations, rather than simply reevaluating its final agent. Each task evaluation can involve multiple LLM calls, repository interactions, and test execution, incurring both runtime and API expenditure. Post-hoc analysis also requires substantial evaluation beyond the main results: The accepted-lineage analysis in Figure~\ref{fig:posthoc_outcomes}(b) likewise requires full 60-task evaluations of intermediate checkpoints. Evaluating every integrated candidate over $T=10$ rounds would require at most $60\times10=600$ rollouts for one backbone on this benchmark. The first-step comparison covers eight agents on 60 tasks (480 agent--task outcomes), while the five-step diagnostic covers 150 outcomes. These analyses examine intermediate and exploratory agents whose task performance is not observed during \method{} evolution. Tables~\ref{tab:elite_run_performance} and~\ref{tab:baseline_repeated_runs} report individual run results and sample standard deviations.

\subsubsection{Additional Experimental Settings and Validity Gates}
\label{app:additional_revision_settings}

\begingroup

\paragraph{Experimental settings.}

\begin{enumerate}[
    label=(\arabic*),
    leftmargin=2em,
    topsep=0.3\baselineskip,
    itemsep=0.5\baselineskip,
    parsep=0.2\baselineskip
]

\item \textbf{Effectiveness across initial scaffolds.}
We use mini-swe-agent v2.4.6 with a fixed
\texttt{Claude-opus-5.5} backbone.
Ten evolution rounds yield six accepted children, numbered by acceptance
order. We evaluate the initial agent and accepted child 6 on Pro,
Verified, and Polyglot using 60, 50, and 50 tasks, respectively.
Accepted children 1 and 3 are also evaluated on the 60 Pro tasks.
The retained children are evaluated post hoc, without feeding their
scores back into evolution.
Table~\ref{tab:mini_swe_agent} reports the results.

\item \textbf{Ablation experiments.}
All variants use the main DeepSeek configuration, initial scaffold,
evaluation tasks, and number of evolution rounds.
Full \method{} and the no-gates ablation use three independent evolution
runs; all other variants use one.
Table~\ref{tab:ablations} reports their scores.
Candidate counts and computation follow each variant's pipeline.

\begin{enumerate}[
    label=\roman*.,
    leftmargin=1.5em,
    topsep=0.3\baselineskip,
    itemsep=0.4\baselineskip,
    parsep=0.2\baselineskip
]

\item \textbf{Alternative selection strategies.}
These variants retain seven independently implemented children and
replace mechanism integration with child selection.
(a) \emph{Select-one (LLM-based)} uses an LLM to choose one child
as the next agent.
(b) \emph{SIFT-style Select-one} uses pairwise LLM judgments aggregated
with a Bradley--Terry model
\citep{fu2026selfimprovementfasttreesearch}.
Both LLM-based strategies operate without downstream candidate scores.
(c) \emph{Evaluation-based Select-one} evaluates all seven children on
the same 30 Pro tasks within each round and selects the highest-scoring
child, requiring $7\times30=210$ child--task evaluations per round.

\item \textbf{Exploration and integration ablations.}
(a) \emph{Single exploration} sets $K=1$ and directly retains the valid
exploratory child as the next agent, removing multi-branch exploration
and integration.
(b) \emph{Direct joint implementation} generates seven improvement
proposals and implements them jointly as one child, omitting the
separate exploratory implementations and their subsequent integration.

\item \textbf{Other component ablations.}
(a) \emph{w/o hypothesis generation} directly proposes self-modifications
without structured hypothesis generation, retaining $K=7$, integration,
and both validity gates.
(b) \emph{w/o deterministic gates} removes both validity gates, allowing
all exploratory children into integration and accepting the integrated
child without deterministic filtering.

\end{enumerate} 
\end{enumerate} 

\paragraph{Validity gates.}
\label{app:gate_implementation}

Both gates use Python scripts that check compilation,
imports, and predefined program assertions.
These checks determine structural and operational validity without
LLM judgments or downstream benchmark scores.
Table~\ref{tab:gate_checks} summarizes their scope and failure handling.
Their coverage concerns basic operability rather than comprehensive
semantic correctness or downstream task performance.

\begin{table}[htbp]
\centering

\small
\setlength{\tabcolsep}{6pt}
\renewcommand{\arraystretch}{1.12}
\caption{\textbf{Deterministic validity gates and failure handling.}}
\label{tab:gate_checks}

\begin{tabularx}{\linewidth}{
    @{}
    >{\raggedright\arraybackslash}p{0.20\linewidth}
    >{\raggedright\arraybackslash}X
    >{\raggedright\arraybackslash}p{0.21\linewidth}
    @{}
}
\toprule
\textbf{Stage}
& \textbf{Main checks}
& \textbf{On failure} \\
\midrule

\textbf{Exploration gate}\par
(After each independent implementation)
&
\begin{itemize}[
    leftmargin=1.1em,
    labelsep=0.4em,
    topsep=0pt,
    partopsep=0pt,
    itemsep=1pt,
    parsep=0pt
]
    \item Nonempty patch.
    \item Compilation of key and modified Python files.
    \item Agent-module imports and tool loading.
    \item Permitted modification scope.
    \item Deployment rules for new dependencies.
\end{itemize}
&
Exclude the exploratory child from integration.
\\
\midrule

\textbf{Acceptance gate}\par
(After integration)
&
\begin{itemize}[
    leftmargin=1.1em,
    labelsep=0.4em,
    topsep=0pt,
    partopsep=0pt,
    itemsep=1pt,
    parsep=0pt
]
    \item Structural and deployment validity.
    \item Smoke tests of the tool-call protocol and main agent flow:
    \begin{itemize}[
        label=\textendash,
        leftmargin=1.1em,
        labelsep=0.4em,
        topsep=1pt,
        partopsep=0pt,
        itemsep=1pt,
        parsep=0pt
    ]
        \item Call/return correspondence.
        \item Message ordering.
        \item Bounded execution in predefined scenarios.
    \end{itemize}
\end{itemize}
&
Reject the integrated child and retain the current parent
for the next generation.
\\

\bottomrule
\end{tabularx}
\end{table}

\endgroup

\begin{algorithm}[htbp]
\caption{\method self-improvement}
\label{alg:elite}
\small
\begin{algorithmic}[1]
\Require initial agent $\mathcal{A}_0$, generations $T$, exploration width $K$
\For{$t=0,\ldots,T-1$}
    \State collect compact evolution history $m_t$
    \State $\mathcal{H}_t \gets \mathcal{P}(\mathcal{A}_t,m_t;K)$
    \State $\mathcal{V}_t \gets \emptyset$
    \For{$k=1,\ldots,K$ \textbf{in parallel}}
        \State $\Delta_t^{(k)} \gets \mathcal{E}(\mathcal{A}_t,h_t^{(k)})$
        \State $\mathcal{A}_t^{(k)} \gets \operatorname{Apply}(\mathcal{A}_t,\Delta_t^{(k)})$
        \If{$G_{\mathrm{exp}}(\mathcal{A}_t^{(k)})=1$}
            \State $\mathcal{V}_t \gets \mathcal{V}_t \cup \{\mathcal{A}_t^{(k)}\}$
        \EndIf
    \EndFor
    \If{$\mathcal{V}_t=\emptyset$}
        \State $\mathcal{A}_{t+1} \gets \mathcal{A}_t$
        \State \textbf{continue}
    \EndIf
    \State construct compact evidence records for children in $\mathcal{V}_t$
    \State $\Delta_t^{\mathrm{int}} \gets \mathcal{I}(\mathcal{A}_t,\mathcal{V}_t)$
    \State $\mathcal{A}_t^{\mathrm{int}} \gets \operatorname{Apply}(\mathcal{A}_t,\Delta_t^{\mathrm{int}})$
    \If{$G_{\mathrm{acc}}(\mathcal{A}_t^{\mathrm{int}})=1$}
        \State $\mathcal{A}_{t+1} \gets \mathcal{A}_t^{\mathrm{int}}$
    \Else
        \State $\mathcal{A}_{t+1} \gets \mathcal{A}_t$
    \EndIf
\EndFor
\State \Return $\mathcal{A}_T$
\Comment{No downstream task evaluation enters the self-improvement loop}
\end{algorithmic}
\end{algorithm}

\subsection{\method Algorithm}
\label{app:elite_algorithm}

Algorithm~\ref{alg:elite} summarizes the complete \method self-improvement procedure. At each generation, valid exploratory children are integrated into a candidate descendant; if no child passes the exploration gate or the integrated descendant fails the acceptance gate, the current agent is retained for the next generation.

\subsection{Mechanism, Scope, and Future Work}
\label{app:mechanism_and_scope}

\paragraph{Implementation-level reasoning.}
The underlying LLM parameters remain fixed, while the scaffold controls tool use, execution feedback, and the problem-solving process. The model's programming knowledge and access to source code allow it to identify concrete limitations, such as missing exit-status feedback, blocking output reads, or incomplete parsing. Executable exploration makes the proposed mechanisms, interfaces, and dependencies explicit, providing evidence for integration to reconcile complementary functionality, redundancy, and conflicts.
For example, the local code probe in Appendix~\ref{app:combined_cases} shows that the integrated implementation combines exit-status feedback with continuous output reading, returning all 100,000 output lines and the correct exit code. Deterministic gates check operational validity, while benchmark evaluations measure the resulting task performance.

\paragraph{Evolution dynamics.}
The observed trajectories show overall gains alongside intermediate variation. In the original DeepSeek run, Pro improves from 81.7\% after the first accepted update to 86.7\% at the end, with temporary declines following two intermediate accepted updates. In the mini-swe-agent run, ten evolution rounds yield six accepted children. Pro increases from 71.7\% initially to 73.3\%, 80.0\%, and 83.3\% for accepted children 1, 3, and 6, respectively (Table~\ref{tab:mini_swe_agent}). These results illustrate improvement over multiple accepted updates. Longer evolution runs would help characterize how progress changes as readily identifiable implementation issues are resolved and remaining bottlenecks depend increasingly on task behavior and interactions among mechanisms.

% \paragraph{Evaluation scope.}
% Our experiments cover two LLM backbones in the main comparison and an additional scaffold--backbone combination using mini-swe-agent. The DeepSeek comparison includes three independent runs per method; Qwen baselines and most ablation variants use one run. Evaluation uses selected benchmark subsets. Additional repetitions and broader task coverage would improve the precision of performance estimates and help characterize how improvement depends on the initial scaffold and the underlying model's programming capabilities.

\paragraph{Evaluation scope.}
Our evaluation focuses on coding tasks, covering two LLM backbones in the main comparison and an additional scaffold--backbone combination using mini-swe-agent. The DeepSeek comparison includes three independent runs per method; Qwen baselines and most ablation variants use one run. Evaluation uses selected benchmark subsets. Additional repetitions and broader task coverage would improve the precision of performance estimates and help characterize how improvement depends on the initial scaffold and the underlying model's programming capabilities. Beyond coding, evaluation could span diverse task domains, interaction environments, and modalities, including long-horizon tool use \citep{li2026tooldecathlonbenchmarkinglanguage}, memory-dependent tasks \citep{he2026memoryarenabenchmarkingagentmemory, nanbo2026historyreplaysitselfunleashing, ren2026gatemembenchmarkingmemorygovernance}, web search \citep{li2026deepresearchbenchiidiagnosing, chen2026trustllmsearchagents}, and multimodal interaction \citep{li2026omnigaianativeomnimodalai}, while assessing safety and robustness across these settings \citep{li2026agentdynagentsecuritydefenses,das2026seabenchbenchmarkingendogenousmisalignment}.

\paragraph{Future work.}
A natural extension is to combine same-parent integration with archive-based search. An archive could retain historical agents, select a parent for exploration and integration, and add the resulting valid descendant while preserving alternative lineages. This would allow the search to revisit intermediate stepping stones and explore multiple evolutionary paths. Developing archive-retention and parent-scheduling strategies based on implementation evidence is a promising direction for extending \method{} while preserving its independence from downstream candidate
task feedback.
\endgroup
\end{document}